\makeatletter\def\graphicscache@inhibit{true}\makeatother
\documentclass[preprint]{elsarticle}

\usepackage{hyperref}

\usepackage[utf8]{inputenc}
\usepackage{lmodern}
\usepackage[T1]{fontenc}
\usepackage{microtype}
\usepackage{textcomp}
\usepackage{graphicx} %

\usepackage{pgfplots}
\pgfplotsset{compat=newest}
\usepackage{subfig}
\usepackage{multirow}
\usepackage{tabularx}
\usepackage{threeparttable}
\usepackage{booktabs}
\usepackage{array}
\usepackage{float}
\usepackage{graphbox}
\usepackage{tikz,tikz-3dplot}
\usetikzlibrary{fit,arrows,arrows.meta,automata,backgrounds,calc,chains,%
decorations.markings,decorations.pathreplacing,decorations.pathmorphing,%
matrix,positioning,shapes,shapes.geometric,shapes.symbols,spy,trees,tikzmark}
\usepackage{balance}
\usepackage[binary-units=true,product-units=single,per-mode=symbol,range-units=single,range-phrase=\,--\,,detect-all]{siunitx}
\DeclareSIUnit\pixel{px}
\usepackage{amsmath} %
\usepackage{amssymb}  %
\usepackage{bm}
\usepackage{acronym}
\usepackage{tablefootnote}

\definecolor{bg_color}{RGB}{95,95,95}

\let\vec\bm

\DeclareMathOperator*{\argmax}{arg\,max}

\DeclareMathOperator*{\normalize}{normalize}
\DeclareMathOperator{\atantwo}{atan2}

\newcommand{\reffig}[1]{Fig.~\ref{#1}}
\newcommand{\reftab}[1]{Tab.~\ref{#1}}
\newcommand{\refsec}[1]{Sec.~\ref{#1}}

\newcommand{\etal}{et al.~}

\newcommand{\wrt}{~w.r.t.~}

\newcommand{\eg}{e.g.,\ }
\newcommand{\cf}{cf.\ }

\usepackage{adjustbox}
\newcolumntype{R}[2]{%
    >{\adjustbox{angle=#1,lap=\width-(#2)}\bgroup}%
    l%
    <{\egroup}%
}
\newcommand*\rot{\multicolumn{1}{R{40}{1em}}}%
\newcolumntype{L}[1]{>{\raggedright\let\newline\\\arraybackslash\hspace{0pt}}m{#1}}

\journal{Robotics and Autonomous Systems}

\begin{document}

\begin{frontmatter}

\title{Real-Time Multi-Modal Semantic Fusion on Unmanned Aerial Vehicles with Label Propagation for Cross-Domain Adaptation}

\author{Simon Bultmann\corref{cor1}}
\ead{bultmann@ais.uni-bonn.de}
\author{Jan Quenzel\corref{}}
\ead{quenzel@ais.uni-bonn.de}
\author{Sven Behnke\corref{}}
\ead{behnke@cs.uni-bonn.de}

\cortext[cor1]{Corresponding author}
\address{Institute for Computer Science VI, Autonomous Intelligent Systems, University of Bonn, Friedrich-Hirzebruch-Allee 8, 53115 Bonn, Germany}

\begin{abstract}
Unmanned aerial vehicles (UAVs) equipped with multiple complementary sensors have tremendous potential for fast autonomous or remote-controlled semantic scene analysis, \eg for disaster examination.

Here, we propose a UAV system for real-time semantic inference and fusion of multiple sensor modalities.
Semantic segmentation of LiDAR scans and RGB images, as well as object detection on RGB and thermal images, run online onboard the UAV computer using lightweight CNN architectures and embedded inference accelerators. We follow a late fusion approach where semantic information from multiple sensor modalities augments 3D point clouds and image segmentation masks while also generating an allocentric semantic map. Label propagation on the semantic map allows for sensor-specific adaptation with cross-modality and cross-domain supervision.

Our system provides augmented semantic images and point clouds with $\approx$\,\SI{9}{\hertz}. We evaluate the integrated system in real-world experiments in an urban  environment and at a disaster test site.
\end{abstract}

\begin{keyword}
robot perception \sep sensor fusion \sep unmanned aerial vehicles \sep semantic segmentation \sep label propagation \sep object detection \sep deep learning
\end{keyword}
\end{frontmatter}

\begin{tikzpicture}[remember picture,overlay]
\node[anchor=north,align=center,font=\sffamily,yshift=-0.5cm] at (current page.north) {%
  Robotics and Autonomous Systems (2022)\\
  DOI: \href{https://doi.org/10.1016/j.robot.2022.104286}{10.1016/j.robot.2022.104286}
};
\end{tikzpicture}%
\vspace{-1.5em}

\section{Introduction}
\label{sec:Introduction}
\begin{figure}[t]
	\centering
	\includegraphics{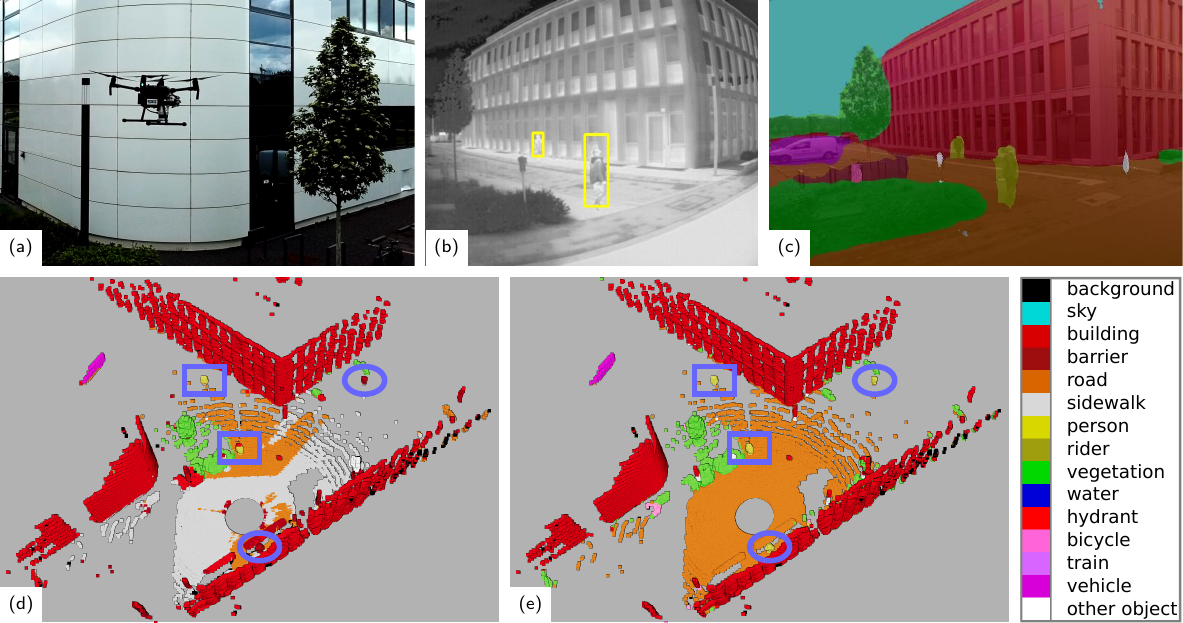}
	\caption{Semantic perception with UAV (a): (b) Person detections in thermal camera, (c) fused image segmentation, and point cloud segmentation (d) before and (e) after label propagation. Persons inside (outside) camera FoV are highlighted with blue rectangles (circles) in (d), (e). Right side: employed semantic classes.}
	\label{fig:teaser}
\end{figure}
Semantic scene understanding is an important prerequisite for solving many tasks with unmanned aerial vehicles (UAVs) or other mobile robots, \eg for disaster examination in search and rescue scenarios~\cite{drz2021ssrr}.
Modern robotic systems employ a multitude of different sensors to perceive their environment, \eg 3D LiDAR, RGB(-D) cameras, and thermal cameras, that capture complementary information about the environment. A LiDAR provides accurate range measurements independent of the lighting conditions, while cameras provide dense texture and color in the visible spectrum. Thermal cameras are especially useful in search and rescue missions as they detect persons or other heat sources regardless of lighting or visibility conditions. 
The combination of all these sensor modalities enables a complete and detailed interpretation of the environment. A semantic map aids inspection tasks~\cite{nguyen_mavnet_2019}, perception-aware path planning~\cite{bartolomei_perception-aware_2020}, and increases robustness and accuracy of simultaneous localization and mapping (SLAM) through the exclusion of dynamic objects during scan matching~\cite{chen_suma_2019}.

In this article, we build upon our recent work~\cite{bultmann2021real}, where we propose a framework for online multi-modal semantic fusion onboard a UAV combining 3D LiDAR range data with 2D color and thermal images. We expand this approach using label propagation for cross-modality supervision to significantly improve LiDAR point cloud segmentation and through additional experiments on a disaster test site. Examples of the semantic perception onboard the UAV are illustrated in \reffig{fig:teaser}.

An embedded inference accelerator and the integrated GPU (iGPU) run inference online, onboard the UAV for mobile, optimized CNN architectures to obtain pixel- resp. pointwise semantic segmentation for RGB images and LiDAR scans, as well as object bounding box detections on RGB and thermal images.
We aggregate extracted semantics for two different output views: A fused segmentation mask for the RGB image which can, \eg be streamed to the operator for direct support of their situation awareness, and a semantically labeled point cloud, providing a 3D semantic scene view which is further integrated into an allocentric map. This late fusion approach is beneficial for multi-rate systems, increasing adaptability to changing sensor configurations and enabling pipelining for efficient hardware usage. The semantic map further allows to adapt specific CNNs to new sensors with unique characteristics using cross-modality supervision from, \eg thermal and color segmentation, through propagating labels via 3D projection.

In summary, our main contributions are:
\begin{itemize}
\item the adaptation of efficient CNN architectures for image and point cloud semantic segmentation and object detection for processing onboard a UAV using embedded inference accelerator and iGPU,
\item the fusion of point cloud, RGB, and thermal modalities into a joint image segmentation mask and a semantically labeled 3D point cloud, 
\item temporal multi-view aggregation of the semantic point cloud and integration into an allocentric map, and
\item evaluation of the proposed integrated system with real-world UAV experiments.
\end{itemize}
In addition to our conference paper~\cite{bultmann2021real} presented at the \textit{2021 European Conference on Mobile Robotics}, we make the following additional contributions:
\begin{itemize}
\item formulating multi-modality fusion in a fully Bayesian manner and improving the clustering of foreground objects, %
\item using label propagation to overcome domain adaptation issues of the LiDAR segmentation network, thereby significantly improving the accuracy of point cloud segmentation, and
\item extended evaluation in large-area UAV flights in an urban environment and on a disaster test site.
\end{itemize}
 
\section{Related Work}
\label{sec:Related_Work}
\paragraph*{Mobile Lightweight Vision CNNs}
Lightweight CNN architectures for computer vision tasks that are efficient and perform well on systems with restricted computational resources, \eg on mobile or embedded platforms, have become of increasing research interest in recent years. The MobileNet architectures~\cite{mobilenetv2_2018,mobilenetv32019} replace classical backbone networks such as ResNets~\cite{he_deep_2016} in many vision models while decreasing the number of parameters and the computational cost significantly, \eg by replacing standard convolutions with depthwise-separable convolutions---at the expense of a slight reduction in accuracy.

In object detection, single-stage architectures such as SSD~\cite{liu_ssd_2016} or YOLO~\cite{Redmon_YOLO_2016} have proven to be efficient in mobile applications through the use of predefined anchors instead of additional region proposal networks. Zhang \etal\cite{Zhang_2019_ICCV} further optimize YOLOv3 for usage onboard a UAV. However, the authors evaluate their network, called Slim-YOLOv3, only on a powerful discrete GPU which is not feasible for integration onboard a typical UAV.

Recently, Xiong \etal introduced MobileDets~\cite{xiong_mobiledets_2021} based on the SSD architecture with MobileNet~v3 backbone and optimized for embedded inference accelerators such as the Google EdgeTPU, which we employ onboard our UAV.

For semantic image segmentation, efficient architectures for inference onboard UAVs have mostly been proposed for specific applications, such as UAV tracking and visual inspection~\cite{nguyen_mavnet_2019} or weed detection for autonomous farming~\cite{weedNet2018}. %
The DeepLab~v3+ architecture~\cite{deeplabv3plus2018} shows state-of-the-art performance on large, general datasets and includes elements of MobileNet architectures such as depth\-wise-separable convolutions for efficient computation. In our work, we employ a DeepLab~v3+ model with MobileNet~v3 backbone for image segmentation.

For point cloud semantic segmentation, projection-based methods~\cite{cortinhal_salsanext_2020,milioto_rangenet_2019,xu_squeezessegv3_2020} utilize the image-like 2D structure of rotating LiDARs. This allows performing efficient 2D-convolutions and using well-known techniques from image segmentation. The downside of this approach is the restriction to single LiDAR scans in contrast to larger aggregated point clouds~\cite{qi2021offboard}. In this work, we adopt the SalsaNext architecture~\cite{cortinhal_salsanext_2020}, trained on the large-scale SemanticKITTI dataset~\cite{behley2019iccv} for autonomous driving, as it shows a good speed-accuracy trade-off.

\paragraph*{Multi-Modal Semantic Fusion}
Mobile robotic systems, such as UAVs or self-driving cars, are often equipped with both camera and LiDAR sensors, as they provide complementary information. A LiDAR accurately measures ranges sparsely and independent of lighting conditions while cameras provide dense textures and colors. Hence, research focused on the fusion of camera and LiDAR for 3D detection and segmentation in the context of autonomous driving.

Xu \etal propose PointFusion~\cite{Xu_pointfusion_2018}, a two-stage pipeline for 3D bounding-box detection. It first processes a LiDAR scan with PointNet~\cite{Qi_2017_pointnet} and an image with ResNet~\cite{he_deep_2016} independently, before fusing them on feature level with an MLP.

Meyer \etal\cite{Meyer_2019_CVPR_Workshops} take a similar sequential feature-level fusion approach, addressing both 3D object detection and dense segmentation. The feature-level fusion requires representing the LiDAR scan as a range image. Range and color image are cropped to the overlapping field-of-view (FoV), reducing the \SI{360}{\degree} horizontal FoV of the LiDAR to only \SI{90}{\degree}.

Vora \etal\cite{Vora_2020_CVPR} propose to in-paint point clouds with image segmentation by projecting LiDAR points into the image and assigning segmentation scores of the pixels. A 3D object detection network then processes the augmented point cloud.

LIF-Seg by Zhao \etal\cite{zhao2021lifseg} improves upon the LiDAR segmentation network Cylinder3D~\cite{zhu2020cylindrical} through early- and middle-fusion with color images.
Image patches around the projected points provide per-point color context for early-fusion, while mid-fusion concatenates semantic features from LiDAR and image, processed with Cylinder3D and DeepLab~v3+, respectively, before processing with an additional refinement sub-network based on Cylinder3D for final semantic labels.

\paragraph*{Semantic Mapping}
Many high-level robotic tasks benefit from or require semantic information about the environment.
For this, semantic mapping systems build an allocentric semantic environment model, anchored in a fixed, global coordinate frame.

SemanticFusion~\cite{mccormac_semanticfusion_2017} models surfaces as surfels where a Gaussian approximates the point distribution. For SLAM, this approach builds on ElasticFusion~\cite{whelan2015elasticfusion} and requires an RGB-D camera. A CNN generates pixel-wise class probabilities from the color image. Their fusion takes a Bayesian approach assuming that individual segmentations are independent and stores all class probabilities per surfel.
Kimera~\cite{rosinol2020kimera} is a modular metric-semantic stereo-inertial-SLAM framework. Its semantic mapping module adopts Voxblox~\cite{oleynikova2017voxblox} to build a truncated signed distance field (TSDF) map of the surface geometry of room-scale indoor environments, integrating semantic segmentation information via a similar Bayesian fusion approach.
Grinvald \etal\cite{grinvald2019vol} represent individual object instances of known and previously unseen classes in the semantic map, providing object-level information for higher-level reasoning.
With Recurrent-OctoMap, Sun \etal\cite{sun_recurrent-octomap_2018} aim at long-term mapping within changing environments. Here, each cell within the OctoMap~\cite{hornung2013octomap} contains an LSTM fusing point-wise semantic features and all LSTMs share weights.

Landgraf \etal\cite{landgraf2020comparing} compare two fusion strategies, first labeling individual views followed by Bayesian fusion versus creating a joint map and labeling it at once. Both strategies show similar results with view-based being more strongly influenced by depth noise while map-based depends on correct poses.

Other works propose alternatives to the probabilistic Bayesian update for fusing semantic labels from multi-view 2D images into a 3D map.
Mascaro \etal\cite{mascaro2021diffuser} build a sparse diffusion graph connecting 2D pixels to 3D points and 3D points to their K nearest neighbors to propagate labels from a 2D image segmentation to the 3D model. After graph construction, iterative multiplication of the label matrix with a probabilistic transition matrix yields the diffused semantic labels.
Berrio \etal\cite{berrio2020proj} use an adapted softmax weighting scheme based on class prevalence within SLIC super pixels to weight individual per-pixel class scores. Motion correction and masking of occluded points are further employed to improve semantic projection accuracy.

For more specific application scenarios, Maturana \etal\cite{maturana_looking_2017} propose to extend existing digital elevation maps (DEM) with the detection of cars from UAVs.
Dengler \etal\cite{dengler2021ecmr} aim for real-time service robotics applications with an object-centric 2D/3D map representation including a 2D polygon of object shape and an object-oriented bounding box in the x-y-plane together with the center of mass and object point cloud. Faster R-CNN~\cite{ren2017faster} detects objects from color images of an RGB-D camera. Euclidean clustering on depth measurements segments small objects geometrically before projection onto the x-y-plane. A refinement step recomputes the biggest clusters after a certain number of fusions to counter errors due to incorrect odometry.
The LiDAR surfel mapping SuMa++ by Chen \etal\cite{chen_suma_2019} uses a surfel's semantic class to further improve the registration accuracy by penalizing inter-class associations during scan matching and surfel update. Here, the projection-based RangeNet++~\cite{milioto_rangenet_2019} provides per-point class probabilities.

Rosu \etal\cite{rosu2020semi} extract a mesh from an aggregated point cloud. The Projection of mesh faces into images enables the transfer from image segmentation to a semantic texture. While projection and fusion happen in real time, the required mesh generation and UV-unwrapping are done in pre-processing. Since only the argmax class is of interest and to meet GPU memory limitations, the sparse texture retains a small number of classes with high probability and discards all others. 
The semantic textured mesh enables label propagation to generate pseudo ground-truth. Retraining the image segmentation network including these pseudo-labels produces more consistent segmentations. While Rosu \etal only improve consistency within one modality, we use propagated labels for sensor-specific adaptation across modalities.

\paragraph*{Domain Adaptation and Label Propagation}
In real-world robotics scenarios, a lack of annotated training data is a major issue.
In recent years, substantial research efforts developed techniques for domain adaptation, that help neural networks to transfer perception skills learned from widely-available standard datasets to application-specific target environments.
This often includes adaptation to other sensors with differing characteristics, such as wavelength, resolution, or FoV, compared to the sensor used for capturing the source dataset.
In this context, label propagation automatically provides annotations for the target domain in a semi-supervised manner, \eg by projecting labels from one sensor modality to another.

A first line of work investigates domain adaptation between different LiDAR sensors and datasets.
Langer \etal\cite{langer2020lidar_dom_transfer} tackle domain transfer between LiDAR sensors with different sampling patterns (i.e. 64-beam vs. 32-beam) by fusing scans from the source data domain and raycasting into the target sensor to obtain transferred training examples. During retraining, weights are shared for source and target and geodesic correlation alignment prevents unwanted domainshift.
Yi \etal\cite{yi2021scvn} perform a similar adaptation between LiDAR sensor types but use a two-stage CNN instead, where at first scene completion obtains a denser canonical point cloud before labeling it in the second stage.
Alonso \etal\cite{alonso2021lidaradapt} examine different data alignment strategies to make different LiDAR datasets more similar and include an alignment loss between source and target dataset based on the KL-divergence.
While these works efficiently handle different LiDAR resolutions, the evaluated sensors have similar FoVs, and datasets all stem from urban driving scenarios. Our work, in contrast, handles more drastic viewpoint changes to aerial UAV perspectives and a LiDAR with different resolution and a significantly larger vertical FoV compared to the source data domain.

Several recent works cope with the limited availability of annotated training data through label propagation. 
Z.~Liu \etal\cite{liu2021otoc} use weak supervision to generate pseudo-labels for 3D data using partitioned super-voxels. A graph relates the super-voxels and propagates pseudo-labels to iteratively train two complementary networks for point segmentation and super-voxel relations.
B.~Liu \etal\cite{liu2019iccv} propagate labels for 2D image data from a small target data domain towards a large unlabeled set with a similarity function pretrained on a source domain.

Most closely related to our approach are methods that apply cross-modal label propagation from 2D images to 3D point clouds.
Piewak \etal\cite{piewak2018lilanet} transfer semantic annotations automatically inferred by an image segmentation CNN from the closest image to point clouds by projection, taking linear ego-motion from wheel odometry into account.
We use a similar approach to automatically obtain labels for point clouds from the image modality, but use a spatio-temporally aggregated 3D semantic map as pseudo-label source, instead of one or multiple individual cameras.

Jaritz \etal\cite{jaritz2019xmuda} present a two-branch network for 3D semantic segmentation. Individual networks compute feature maps for LiDAR and camera before retaining only those features at valid projected points in the camera FoV. In parallel to the concatenation of both feature maps before a fused segmentation head, each branch performs single-modality segmentation. During training, the single heads should mimic the fused output by minimizing the cross-modal loss based on KL-divergence. This requires labels for both modalities within the source domain. After initial adaptation to the target data domain, \eg a different dataset without labels, the generation of pseudo-labels in the target data domain and retraining from scratch provides further improvements.
Similarly, Wang \etal\cite{wang2021multistage} use a two-branch network for 3D bounding box detection from LiDAR and images. A gated adaptive fusion subnetwork introduces point-wise projected image features into the LiDAR branch on every layer within the feature encoder. A KL-divergence loss regularizes class predictions between the image and LiDAR branch.
Due to the close coupling between image and LiDAR modalities, the above two methods only use 3D points inside the camera FoV. Our method, on the other hand, segments all LiDAR points in the complete \SI{360}{\degree} horizontal FoV.

In our work, different networks process LiDAR scan, RGB, and thermal images individually. We adopt a pro\-jec\-tion-based approach similar to~\cite{Vora_2020_CVPR} for multi-modal fusion in a multi-rate system. When multiple modalities are available, we merge class probabilities from different sensors using Bayesian fusion. Our mapping integrates augmented point clouds in a sparse voxel hash-map with per voxel full class probabilities. We adapt the Bayesian fusion of SemanticFusion~\cite{mccormac_semanticfusion_2017} to logarithmic form for higher precision and greater numerical stability.

While being less popular in recent work, such a late fusion approach has important practical advantages for deployment on an integrated robotic system. Different FoVs and data rates are easy to handle and intermediate results, such as image segmentation or detections, are useful as stand-alone outputs.
Pipelining also allows for reducing the latency of sequentially executed individual networks during online operation. Furthermore, the smaller, simpler standard architectures of individual networks are easier to adapt and optimize for the embedded inference accelerators employed in this work.

As our target data domain of UAV aerial perspectives with large vertical FoV differs significantly from available large-scale training datasets from autonomous driving scenarios with different viewpoints and LiDAR sensors more focused towards the ground, we use label propagation to retrain the point cloud segmentation CNN with supervision for the target environment.
Pseudo-labels are automatically obtained via 3D projection from RGB and thermal camera modalities, spatio-temporally aggregated in a semantic map to compensate for their narrower FoV\wrt the LiDAR scanner.
Through retraining with this cross-domain supervision, point cloud segmentation is significantly improved, achieving higher mIoU scores on our dataset than the image segmentation used as pseudo-label source, and generalizing to the full LiDAR FoV.
 
\section{Our Method}
\label{sec:method}
\subsection{System Setup}
\begin{figure}[t]
  \centering
  \includegraphics[width=\linewidth]{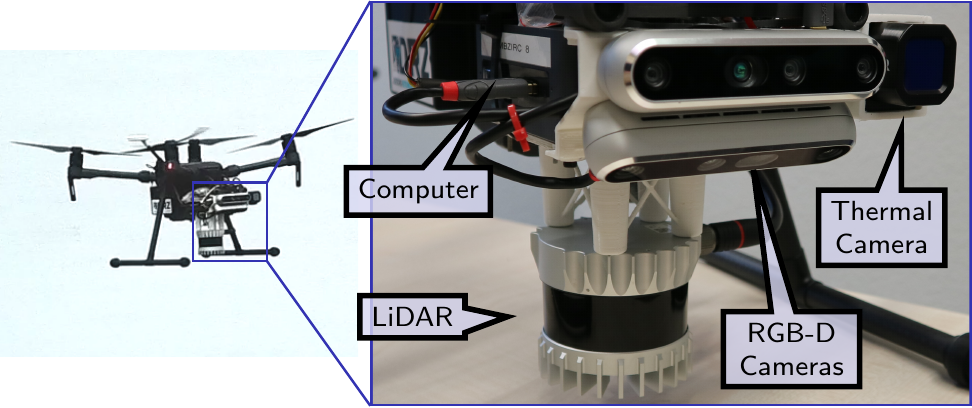}
  \caption{UAV system setup and hardware design.}
  \label{fig:hardware}
\end{figure}
An overview of our UAV system, based on the commercially available DJI Matrice 210 v2 platform, is shown in \reffig{fig:hardware}.
We use an Intel Bean Canyon NUC8i7BEH with a Core i7-8559U processor and \SI{32}{\giga\byte} of RAM as the onboard computer. A Google EdgeTPU connects to the NUC over USB 3.0 and accelerates CNN inference together with the Intel Iris Plus Graphics 655 iGPU of the main processor.
An Ouster OS0-128 3D-LiDAR\footnote{\url{https://ouster.com/products/scanning-lidar/os0-sensor/}} with 128 beams, \SI{360}{\degree} horizontal, and \SI{90}{\degree} vertical opening angles provides range measurements for 3D perception and odometry.
For visual perception, our UAV additionally carries two Intel RealSense D455 RGB-D cameras, mounted on top of each other to increase the vertical field-of-view, and a FLIR ADK thermal camera for, \eg person detection in search and rescue scenarios.
\begin{figure}[t]
  \centering
  \includegraphics[width=\linewidth]{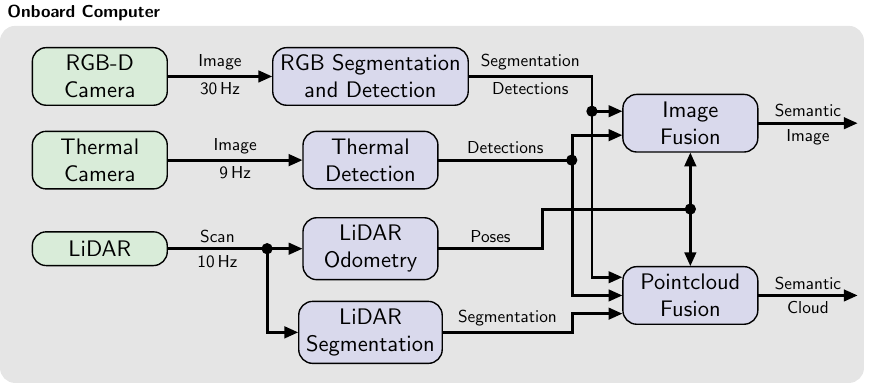}
  \caption{Online perception system overview.}
  \label{fig:system}
\end{figure}
\subsection{Semantic Perception}
An overview of the proposed architecture for multi-modal semantic perception is given in \reffig{fig:system}. We detail individual components in the following.
\subsubsection*{Image Segmentation}
We employ the DeepLabv~3+~\cite{deeplabv3plus2018} architecture with MobileNet~v3~\cite{mobilenetv32019} backbone optimized for Google EdgeTPU Accelerator for semantic segmentation. We train the model on the Mapillary Vistas Dataset~\cite{MVD2017}, reducing the labels to the 15 most relevant classes for the envisaged UAV tasks (\cf\reffig{fig:teaser}). We use an input image size of 849$\,\times\,$481~px during inference, fitting the 16:9 aspect ratio of our camera.
\subsubsection*{Object Detection}
The recent MobileDet architecture~\cite{xiong_mobiledets_2021} is the basis for our object detection. We train the RGB detector on the COCO dataset~\cite{lin_coco_2014} for \textit{person}, \textit{vehicle}, and \textit{bicycle} classes with an input resolution of 848$\,\times\,$480~px. The thermal object detector uses the same architecture taking one-channel 8-bit gray-scale thermal images at the full camera resolution of 640$\,\times\,$512~px as input.
We enable automatic gain correction (AGC) for the thermal camera which adapts and scales the 16-bit raw images to 8-bit, exploiting the full 8-bit value range to provide contrast-rich images. The network is trained on the FLIR ADAS dataset~\cite{flir_dataset}, recorded with the previous generation of our employed sensor in autonomous driving scenarios, with annotations for \textit{persons}, \textit{vehicles}, and \textit{bicycles}.
\subsubsection*{Point Cloud Segmentation}
We adopt the projection-based SalsaNext architecture~\cite{cortinhal_salsanext_2020} taking advantage of the image-like structure of LiDAR measurements. The network is pretrained on the large-scale SemanticKITTI~\cite{behley2019iccv} dataset. The OS0 LiDAR sensor provides measurements at a resolution of 1024$\,\times\,$128.
We compare using the full sensor resolution to subsampling the scans by a factor of two in both vertical and horizontal directions, leading to a network input resolution of 512$\,\times\,$64. Subsampling enables real-time inference on our hardware.
The input channels are range, $x$-, $y$-, $z$-coordinate, and intensity, normalized with the mean and standard deviation of the training dataset. Our LiDAR has a significantly larger vertical FoV of \SI{90}{\degree} compared to the \SI{26.9}{\degree} opening angle of the Velodyne HDL-64E sensor employed in the SemanticKITTI dataset.
The HDL-64E mostly measures downward from the horizontal plane, thus seldomly measuring treetops or other higher obstacles. A different laser wavelength also changes the characteristics of intensity and reflections.
Hence, we adjust the normalization parameters for $z$-coordinate and intensity to facilitate the cross-domain adaptation between training and observed data according to the statistics of the test data captured with our sensor setup, taking up the idea of input data distribution alignment from~\cite{alonso2021lidaradapt}. The $x$- and $y$-coordinate normalization parameters remain the same, as the horizontal field-of-view is identical (\SI{360}{\degree}) for both sensors.
\reffig{fig:pointcloud_semantics} highlights improvements of the segmentation results through the adaptation of the normalization parameters. The point cloud segmentation nonetheless remains noisier and less detailed than the image segmentation.
To further overcome the domain adaptation issues, we retrain the point cloud segmentation network using label propagation (\cf\refsec{sec:labelprop}).
\subsubsection*{Inference Accelerators}
We run the CNN model inference on two different accelerators onboard the UAV PC: The Google EdgeTPU\footnote{\url{https://coral.ai/docs/accelerator/datasheet}}, attached as an external USB device, and the integrated GPU (iGPU) included in most modern processors which is otherwise unused in our system.
The EdgeTPU supports network inference via TensorFlow-lite~\cite{abadi2016tensorflow} and requires quantization of the network weights and activations to 8-bit~\cite{quantization_2018}. The iGPU supports inference via the Intel OpenVINO framework\footnote{\url{https://docs.openvinotoolkit.org/}} in 16- or 32-bit floating-point precision.
\subsection{Multi-Modality Fusion}
\label{sec:mm_fusion}
We adopt a projection-based approach to fuse semantic class scores from image and point cloud CNNs into the semantically labeled output cloud.
Projection onto the image plane requires the transformation of LiDAR points into the respective camera coordinate frame. As LiDAR and cameras operate with different frame rates, the motion between the respective capture times has to be taken into account. The full transformation chain $\vec{T}$ from LiDAR to camera frame is:
\begin{align}
\vec{T} &= {}^{\text{cam}}\vec{T}_{\text{base}} {}^{\text{base}_{t_c}}\vec{T}_{\text{base}_{t_l}} {}^{\text{base}}\vec{T}_{\text{LiDAR}}\,,
\end{align}
using the continuous-time trajectory of the UAV base frame estimated by the LiDAR odometry. Thus, the transformation chain models perspective changes between LiDAR and camera that occur due to dynamic UAV motions.

Bilinear interpolation at the projected point location gives the semantic class scores $\vec{c}_{\text{img}} \in\mathbb{R}^C$ from image segmentation.
We apply the soft-max operation to approximate a normalized probability distribution over all $C=15$ classes used in this work (\cf \reffig{fig:teaser}):
\begin{align}
p_i &= \sigma\left( c_i \right) = \frac{\exp{c_i}}{\sum^C_{j=1}\exp{c_j}}\,,
\end{align}
 obtaining $\vec{p}_\text{img}\in\mathbb{R}^{C}$, with $p_i \in [0,1]$ and $\sum_i p_i = 1$.
Similarly, the application of soft-max to the output of the point cloud CNN for a LiDAR point gives the LiDAR segmentation probability $\vec{p}_\text{LiDAR}$. The Bayesian update rule~\cite{mccormac_semanticfusion_2017} allows to compute the fused class probability under the assumption of independence between sensor modalities:
\begin{align}
\vec{p}_\text{fused} &= \frac{\vec{p}_\text{img} \circ \vec{p}_\text{LiDAR}}{\sum_{i=1}^C p_{i,\text{img}}\,p_{i,\text{LiDAR}}}\,,
\end{align}
with $\circ$ being the coefficient-wise product. 
For better numerical stability, we implement the Bayesian fusion in logarithmic space, as detailed in \refsec{sec:semantic_mapping}.

Furthermore, if a projected point falls inside a detection box in either thermal or color images, the detected class is included in the result. We base the detection probability $p_\text{det}$ on the detector score multiplied with a Gaussian factor with mean at the bounding box center and standard deviation of half the bounding box width resp. height, to reduce unwanted border effects for non-rectangular or non-axis-aligned objects.
As the detector only outputs a score for the detected class, we reconstruct the full probability distribution $\vec{p}_\text{det}$ following the maximum entropy principle: The remaining probability mass $1-p_\text{det}$ is equally distributed over the remaining $C-1$ classes.
Again, both estimates are fused using Bayesian update:
\begin{align}
\vec{p}_\text{fused\_det} &= \frac{\vec{p}_\text{fused} \circ \vec{p}_\text{det}}{\sum_{i=1}^C p_{i,\text{fused}}\,p_{i,\text{det}}}\,.
\end{align}

\begin{figure}
	\centering
	\includegraphics[width=\linewidth]{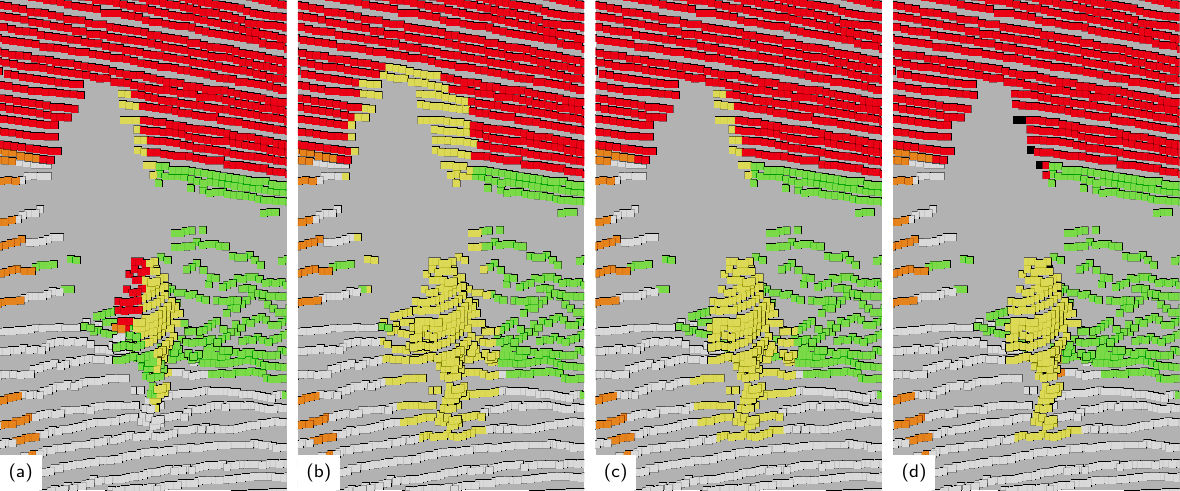}
	\caption{Person segmentation included into point cloud fusing (a) image segmentation only, (b) additionally detection bounding boxes, (c) clustering foreground points within the detection bounding boxes via depth threshold, and (d) Euclidean clustering. The initial segmentation (a), is incomplete and slightly misaligned. Naive bounding box fusion (b) creates many false positives in the background and on the floor. With depth threshold clustering (c), the person is completely segmented without adding many misclassified points in the background. With Euclidean clustering (d), the person is completely segmented and misclassified points in the background and on the floor are reset to their original label from LiDAR segmentation.}
	\label{fig:pointcloud_semantics_person}
\end{figure}
Side-effects of the rectangular detection bounding boxes have to be handled before detection fusion, however, as illustrated in \reffig{fig:pointcloud_semantics_person} with the example of fusing person detections from the RGB or thermal image into the point cloud.
Simple projection of all points into the bounding box will falsely label points in the background as the detected class (\cf \reffig{fig:pointcloud_semantics_person} (b)). To alleviate this issue, points are clustered\wrt their distance in the camera frame before detection fusion.
In our previous work~\cite{bultmann2021real}, we included only points within a fixed threshold of the \SI{25}{\percent} quantile of distances per cluster to focus on foreground objects (\cf \reffig{fig:pointcloud_semantics_person} (c)).
We extend this approach to Euclidean point clustering with an adaptive cluster tolerance threshold $\tau_\text{cluster}$. This yields a more accurate segmentation and better generalizes to different object sizes and distances.
Starting from the seed point at \SI{25}{\percent} quantile distance $d_\text{seed}$, bounding box points within the distance $\tau_\text{cluster}$ are recursively added to the cluster. The cluster tolerance is proportional to $d_\text{seed}$ and the angle increment between two adjacent scan lines, adapting to the LiDAR spatial resolution which covers a vertical FoV of \SI{90}{\degree} with 128 lines: %
\begin{align}
\tau_\text{cluster} = s \cdot d_\text{seed} \cdot \frac{\text{FoV}_\text{vert}}{\text{Res}_\text{vert}} = s \cdot d_\text{seed} \cdot \frac{\pi}{2 \cdot 128}\,,\label{eq:cluster}
\end{align}
where $s=1.5$ is a tolerance factor set to yield complete foreground clusters without adding points on the floor or background (\cf \reffig{fig:pointcloud_semantics_person} (d)). We assume that there is only one valid cluster per bounding box. Only the clustered points are included into detection fusion.
Furthermore, points not added to the cluster that are erroneously labeled as the cluster class, are reset to their original class probability from LiDAR segmentation to correct for border effects in the previous fusion stage (\cf background in \reffig{fig:pointcloud_semantics_person} (a) vs. (d)).
The final segmented point cloud includes the full class probability vector and the argmax class color per point.

We proceed similarly for fusing the initial image segmentation and detections from RGB and thermal cameras into the output semantic image and additionally apply temporal smoothing.
The RGB-D depth enables projection from RGB to thermal image and temporal smoothing provides a more coherent fused segmentation. For temporal fusion, we project the previous image at time $t-1$ with its depth into the current frame at time $t$ and perform exponential smoothing:
\begin{align}
\vec{p}_{\text{smoothed\_img}_{t}} &= \normalize\left(\vec{\alpha} \circ \vec{p}_{\text{img}_{t}} + \left(\vec{1} - \vec{\alpha}\right) \circ \vec{p}_{\text{fused\_img}_{t-1}}\right)\,,\label{eq:tmp_smoothing}\\
\vec{p}_{\text{fused\_img}_t} &= \frac{\vec{p}_{\text{smoothed\_img}_{t}} \circ \vec{p}_{\text{det}_t}}{\sum_{i=1}^C p_{i,\text{smoothed\_img}_{t}}\,p_{i,\text{det}_t}}\,.\label{eq:det_fusion_img}
\end{align}

The smoothing weights $\vec{\alpha}$ differ between the individual semantic classes. For (potentially) dynamic foreground objects, such as \textit{persons} and \textit{vehicles}, less smoothing is applied than for static structures such as buildings and roads. We chose $\alpha_\text{dyn}=0.80$ for dynamic object classes and $\alpha_\text{stat} = 0.25$ for static background classes in our experiments.
The higher $\alpha$-coefficients for dynamic objects make the fused segmentation mask correctly follow dynamic foreground objects moving over image areas that were previously segmented as a background class, as the current frame's segmentation more directly influences the fused output for these semantic classes. The smoothing for background classes significantly reduces temporal jitter in the segmentation\wrt the initial CNN output.
The temporal smoothing~\eqref{eq:tmp_smoothing} can only be applied to pixels with valid depth measurements that have a corresponding projected point from the previous image. We directly use the segmentation class probabilities from the current frame $\vec{p}_{\text{img}_{t}}$ for fusion with the detections in~\eqref{eq:det_fusion_img} otherwise.

\subsection{Semantic Mapping}
\label{sec:semantic_mapping}
MARS LiDAR Odometry~\cite{quenzel2021mars} provides poses to integrate all augmented point clouds within a common map. A uniform grid subdivides the space into cubic volume elements (voxels). Since a dense voxel grid may require a prohibitively large amount of memory although only sparse access occurs, we use sparse voxel hashing. Each voxel fuses all points in its vicinity probabilistically. Additionally, we compute the mean position.
Our fusion scheme follows the reasoning of SemanticFusion~\cite{mccormac_semanticfusion_2017} to use Bayes' Rule assuming independence between semantic segmentations $P\left( l_i \lvert X_k\right)$ for the augmented point cloud $X_k$ with label $l_i$ for class $i$:
\begin{align}
P( l_i \lvert X_{1:k} ) &=  \frac{P\left(l_i \lvert X_{1:k-1}\right)  P \left( l_i \lvert X_k \right)}{\sum_i P\left( l_i \vert X_{1:k-1} \right) P\left( l_i \vert X_k\right)}.\label{eq:bayes}
\end{align}
A naive implementation, as in SemanticFusion, suffers from numerical instability due to the finite precision of the multiplication result. In practice, this leads to all class probabilities being close to zero, \eg when $P\left( l_i \lvert X_k\right)\approx 1$, $P\left( l_j \lvert X_{k+1}\right)\approx 1$ and $P\left( l_i \lvert X_{k+1}\right)\approx 0$, $P\left( l_j \lvert X_k\right)\approx 0$ both class-wise products will be almost zero. This results in a loss of information even after the application of the normalization term and needs continuous reinitialization.

Hence, we switch to log probabilities:
\begin{align}
L_{i,1:k} &= \log\left(P\left(l_i \lvert X_{1:k}\right)\right),\\ 
L_{i,1:k-1} &= \log\left(P\left(l_i \lvert X_{1:k-1}\right)\right),\\
L_{i,k} &= \log\left(P \left( l_i \lvert X_k \right)\right),\\
S_{1:k} &= \log\left({\sum_i P\left( l_i \vert X_{1:k-1} \right) P\left( l_i \vert X_k\right)}\right),
\end{align}
and voxels now store $L_{i,1:k}$ instead of $P\left( l_i \lvert X_{1:k}\right)$.
We obtain for \eqref{eq:bayes} in log form:
\begin{align}
L_{i,1:k} &= S_{i,1:k} - S_{1:k}\,,\\
S_{i,1:k} &= L_{i,1:k-1}+L_{i,k}\,,
\end{align}
while we make use of the following logarithm identity for $S_{1:k}$ to factorize out the largest summand with index $m = \argmax_i(x_i)$ for numerical stability:
\begin{align}
\log\left(\sum_i x_i\right) &= \log\left( x_m \right) + \log \left(1 + \sum_{i \neq m}\frac{x_i}{x_m}\right),\\
&= \log\left( x_m \right) + \log \left(1 + \sum_{i \neq m} \exp^{\log(x_i)-\log(x_m)}\right).
\end{align}
Thus, we compute $S_{1:k}$ as follows:
\begin{align}
S_{1:k} &= S_{m,1:k} + \log\left(1+\sum_{i\neq m}{\exp^{S_{i,1:k}-S_{m,1:k}}}\right).
\end{align}

Ideally, we would directly fuse network outputs before soft-max to save additional $\exp$ and $\log$ evaluations, but since the individual outputs may be arbitrarily scaled, this step is necessary.

An infinite time horizon of the semantic map, fusing all scans, may not be necessary or wanted---depending on the use-case, e.g. for global vs. local planning. Hence, we employ a fixed-size double-ended queue (deque) per voxel for a shorter time horizon of $n$ scans that merges all points per scan. Fusion of per-scan log probabilities yields the voxels' class probabilities. Older scans are either removed completely or fused into the infinite time horizon estimate.

\subsection{Label Propagation}
\label{sec:labelprop}
The employed LiDAR segmentation CNN, pretrained on the SemanticKITTI dataset~\cite{behley2019iccv}, shows limited performance in our test scenarios (\cf \refsec{sec:eval_outdoor}). This is due to cross-domain adaptation issues between training and observed data as the UAV employs a LiDAR sensor different from SemanticKITTI, with different vertical FoV, laser wavelength, and optical system. To the best of our knowledge, no large-scale semantically annotated training datasets are available using the employed Ouster OS0-128 sensor.

To overcome these issues, we retrain the CNN using our sensor's FoV parameters by (1) complementing the SemanticKITTI training data with the recently published Paris-CARLA-3D dataset~\cite{deschaud2021paris} and (2) automatically generating pseudo-labels for cross-modal supervision from the fused semantics of RGB and thermal camera from outdoor flights with our UAV system.

The Paris-CARLA-3D dataset contains aggregated point clouds from three streets in Paris over about \SI{550}{\meter} linear distance and a Velodyne HDL32 LiDAR sensor similar to the one used for SemanticKITTI. However, the sensor was tilted, allowing high structures such as buildings to be fully mapped. We only utilize the real-world part of the dataset.
To obtain labeled single scans to complement the training data, we project points from the aggregated cloud into simulated viewpoints with the characteristics of the Ouster OS0 LiDAR at positions following the original vehicle trajectory from the dataset, but at larger height, further adapting to our UAV use case.
The projection of a LiDAR point $(x,y,z)^\intercal$ in the sensor frame to image coordinates $(u,v)^\intercal$ is given as in~\cite{cortinhal_salsanext_2020} by:
\begin{align}
\begin{pmatrix}u\\v\end{pmatrix} &= \begin{pmatrix}0.5 \left(1 - \atantwo\left(y,x\right)\pi^{-1}\right)w \\\left(1 - \left(\arcsin\left(zr^{-1}\right) + f_\text{down}\right)f^{-1}\right)h\end{pmatrix}\,,
\end{align}
with $h$, $w$ denoting the height, resp. width of the projected image, $r = \sqrt{x^2+y^2+z^2}$ the range of each point and $f = \left|f_\text{down}\right| + |f_\text{up}|$ the vertical field-of-view ($f_\text{down} = f_\text{up} = \SI{45}{\degree}$ for OS0).
When multiple points are projected to the same image coordinates, only the closest one is retained. The maximum range is set to \SI{50}{\meter}.

\begin{figure}[t]
	\centering
	\includegraphics[width=\linewidth]{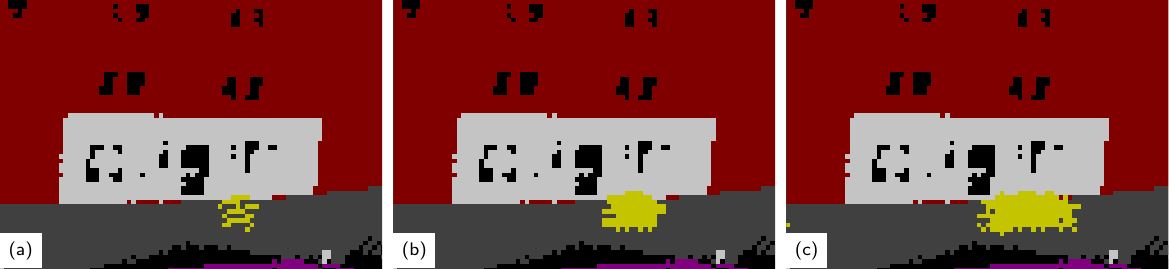}
	\caption{Paris-CARLA-3D dataset: Projection of dynamic persons (yellow) into the virtual OS0 LiDAR view using dataset scans from (a) only the current, (b) $\pm 2$, and (c) $\pm 5$ adjacent positions. The foreground person is incomplete when using just a single scan (a) but significant movement artifacts appear for a larger scan window (c). We use a window of $\pm 2$ (b) for our experiments as compromise between complete scans and remaining artifacts.}
	\label{fig:paris_sampling}
\end{figure}
To avoid artifacts from dynamic objects (e.g. cars and persons), points of these semantic classes are only projected from a limited number of scans captured close to the current simulated position. Here, a compromise must be made between complete, dense scans and remaining artifacts as illustrated in \reffig{fig:paris_sampling}. We choose a window of $\pm 2$ scans for our experiments.

\begin{figure}[t]
	\centering
	\includegraphics[width=\linewidth]{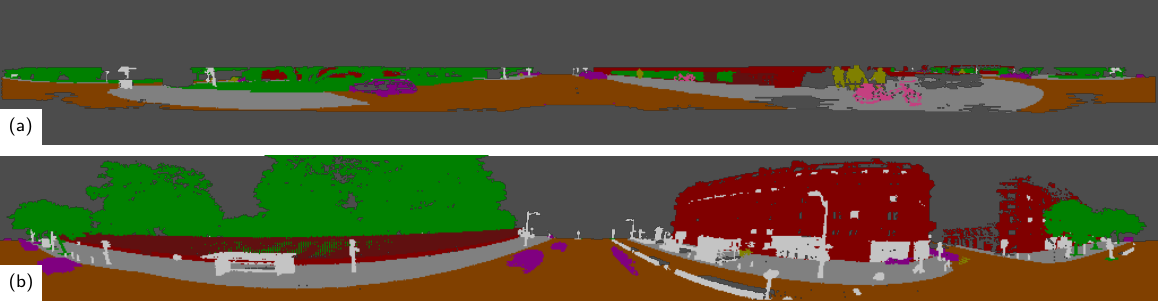}
	\caption{Point cloud labels projected into the FoV of the employed Ouster OS0 LiDAR for sample scenes from (a) SemanticKITTI~\cite{behley2019iccv} and (b) Paris-CARLA-3D~\cite{deschaud2021paris} datasets. SemanticKITTI covers a significantly smaller vertical FoV than our sensor, while the aggregated cloud from Paris-CARLA-3D covers the entire FoV.}
	\label{fig:training_data}
\end{figure}
The differences between the two datasets are illustrated in \reffig{fig:training_data} where labels from SemanticKITTI and Paris-CARLA-3D are shown projected in the Ouster OS0 sensor FoV. The SemanticKITTI data covers only a small part of the vertical FoV with the top of buildings rarely visible, while the scans obtained from Paris-CARLA-3D cover the full FoV and show complete building structures, similar to the data observed by the UAV system.
Yet, the variability of Paris-CARLA-3D (\SI{550}{\meter} distance, single city area) is limited compared to SemanticKITTI (\SI{39.2}{\kilo\meter} distance, different urban, rural, and highway areas~\cite{geiger2012cvpr}), and some artifacts from moving objects remain.
Furthermore, as a different LiDAR sensor is used, the intensity channel, computed from the magnitude of laser reflection, is still only partially comparable to the Ouster OS0 which uses a different wavelength and optical system.

\begin{figure}[t]
  \centering
  \includegraphics[width=\linewidth]{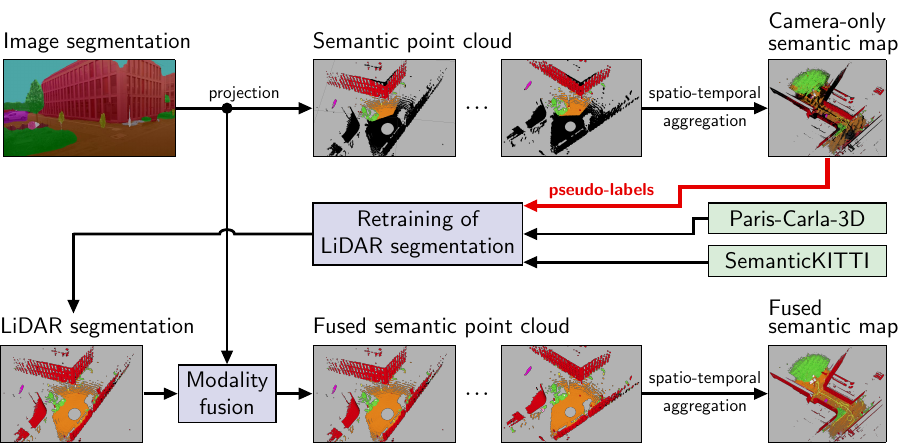}
  \caption{Overview of label propagation approach: Image segmentation is projected on point cloud and aggregated into camera-only semantic map, used as source for pseudo labels. LiDAR segmentation is retrained combining own data with pseudo-labels, Paris-CARLA-3D, and SemanticKITTI. Resulting semantic point clouds, fusing image and retrained point cloud segmentation, are completely and accurately annotated.}
  \label{fig:labelprop_approach}
\end{figure}

To obtain training data from the actual sensor, we use label propagation for cross-modal domain adaptation to automatically generate pseudo-labels for data captured during flights with our UAV system.
An overview of the proposed approach is given in \reffig{fig:labelprop_approach}.

\begin{figure}[t]
	\centering
	\includegraphics[width=\linewidth]{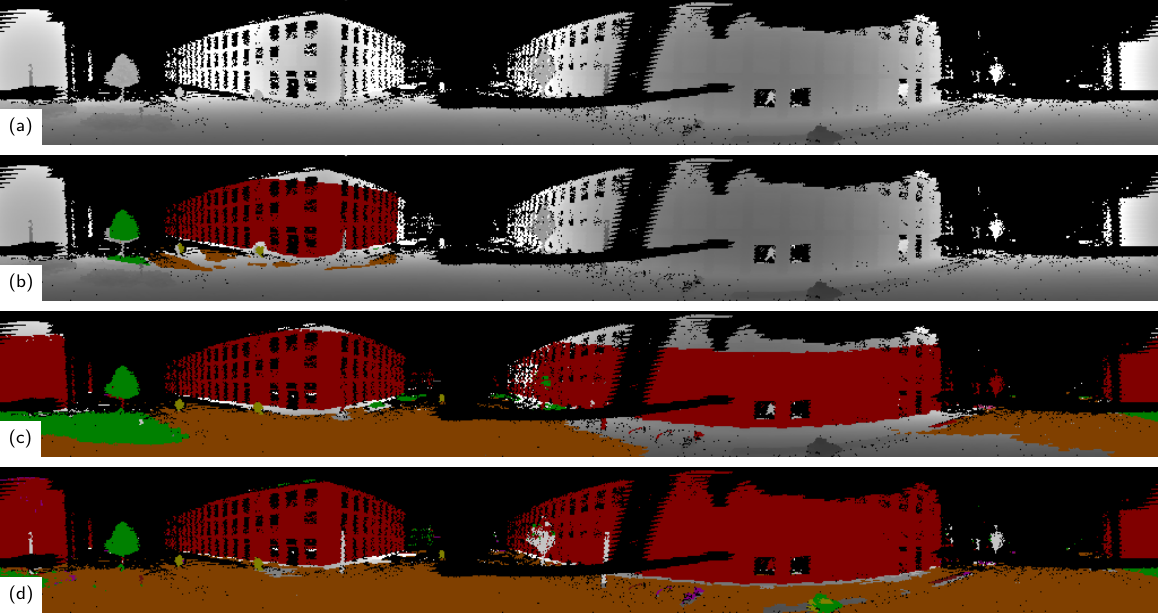}
	\caption{Label propagation using cross-modal supervision: (a) range channel of projected scan, (b) pseudo-labels from single camera overlay, (c) from aggregated camera-only map, and (d) from aggregated fused map (\cf \reffig{fig:semantic_map}~(b, c)). Unlabeled areas are depicted in gray\-scale. UAV legs lead to areas without valid measurements (black) at close range. Single camera overlay provides supervision only for a small part of the scan. Supervision from aggregated maps is more complete and the camera-only map~(c) is more accurate than the fused map~(d).}
	\label{fig:labelprop}
\end{figure}

As the semantic information from the RGB and thermal camera modalities is significantly more reliable than the initial point cloud segmentation (\cf \refsec{sec:eval_outdoor}), we only use the camera semantics as pseudo-label source.
For this, we fuse the RGB and thermal camera semantic information into the point cloud, without including the outputs of the LiDAR segmentation CNN. The obtained pseudo-labels are illustrated in \reffig{fig:labelprop}.
As the FoV of the cameras is significantly smaller than that of the LiDAR, only a small part of each individual scan can be labeled with this cross-modal supervision. However, after aggregation over the complete flight, the semantic map can provide pseudo-labels for almost the complete scan. Only the operator of the UAV (person to the bottom right of \reffig{fig:labelprop}) is not annotated as they always stayed behind the UAV and never were in the camera FoV.
For reference, we also compare to using pseudo-labels from the map aggregated from fused semantic clouds. We find, however, that this supervision is too imprecise to achieve significant improvements (\cf \reftab{tab:iou_data}) since the noisier raw point cloud segmentation is included (\eg wrongly labeled vegetation to the left and person to the bottom right of \reffig{fig:labelprop}~(d)). It is crucial that the pseudo-labels used for re-training are as accurate as possible and it is better to leave uncertain parts unlabeled than to fill them with imprecise labels.

The semantic map that serves as source for pseudo-labels was generated fully automatically from captured data, using the proposed system for multi-modality fusion and semantic mapping, without any manual supervision.
Additionally, a single, automated post-processing step improves the label quality: Points on the ground plane that are not labeled as a ground class (i.e. \textit{road}, \textit{sidewalk}, or \textit{vegetation}) are reset to the unknown semantic class, to correct for unwanted artifacts at object borders.
Similar to Rosu \etal\cite{rosu2020semi}, we use only high-confidence pseudo-labels (minimum confidence of \SI{80}{\percent}) leading to unlabeled gray regions between areas of different semantic classes in \reffig{fig:labelprop}~(b). After aggregation of multiple viewpoints in the semantic map, most labels have confidence close to $1$ and the borders between semantic classes are sharp.

For retraining the point cloud segmentation network, we complement the SemanticKITTI training data with scans obtained from Paris-CARLA-3D and self-recorded UAV flights with pseudo-labels. The amount of additional data is chosen to be comparable to the original $\approx$ 20k scans of SemanticKITTI, as proposed by 
Rosu \etal\cite{rosu2020semi}. We increase the batch size to $64$ to compensate for the lower signal-to-noise ratio due to the noisier pseudo-labels.
Furthermore, we increase the magnitude of the data augmentation transformations\wrt original SalsaNext training parameters~\cite{cortinhal_salsanext_2020}, to account for the lower variability of the additional scenes.
After retraining, we compare networks using the five original input channels (range, $x$-, $y$-, $z$-coordinate, and intensity) against ones not using intensity information, as this channel is the most difficult to adapt for between changing sensor types.
If not stated otherwise, the network is retrained with an input resolution of 512$\,\times\,$64, using SemanticKITTI, Paris-CARLA-3D, and scans from our own data collection with pseudo-labels from the camera-only semantic map. Different parameters are compared in the ablation studies in \refsec{sec:Evaluation}.
 
\section{Evaluation}
\label{sec:Evaluation}
We first evaluate inference speed and computational efficiency of the employed CNN models and then show results from outdoor UAV flights in an urban environment and on a disaster test site.

\begin{figure}[t]
  \centering
  \includegraphics[width=\linewidth]{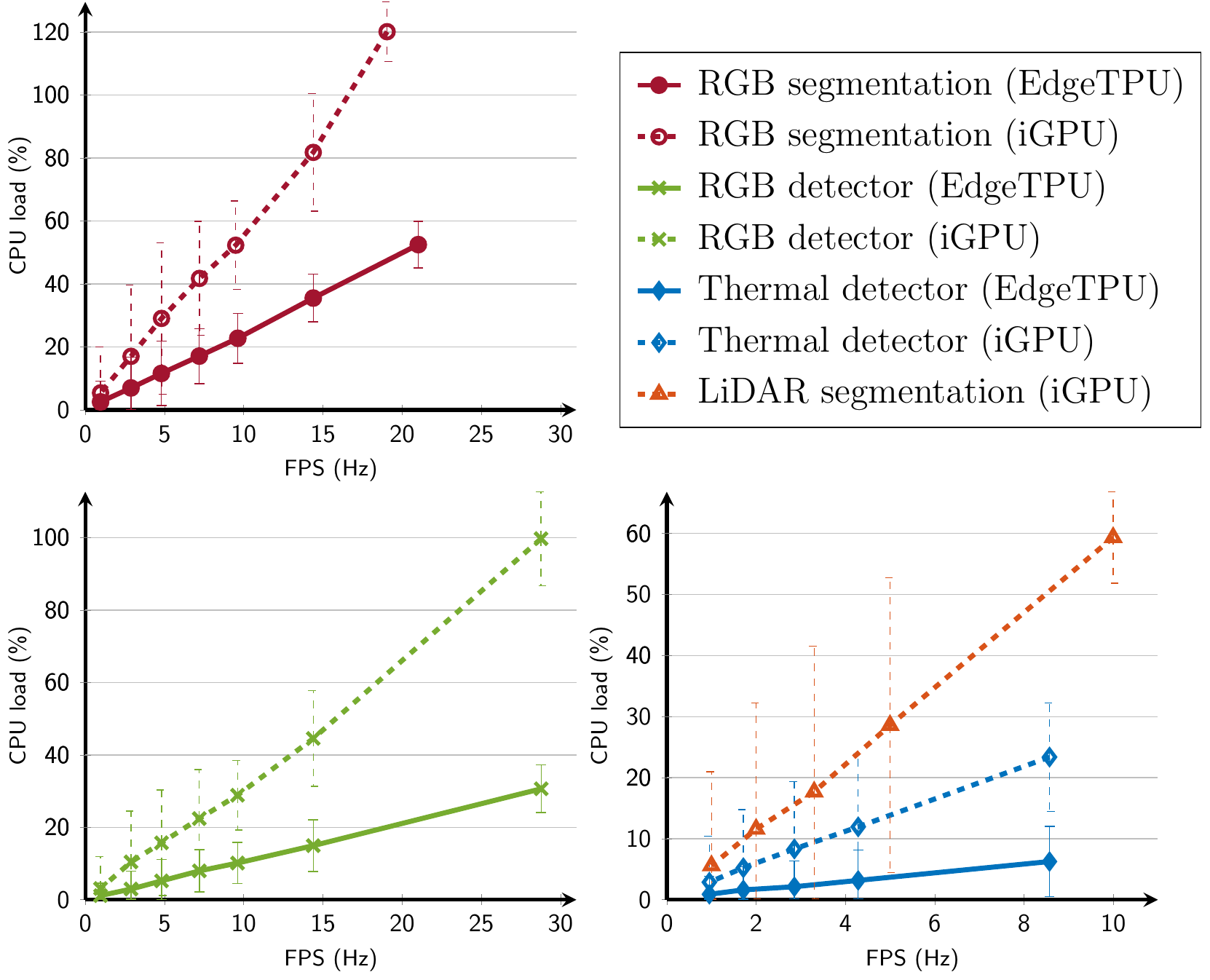}
  \caption{CPU load of the CNN inference of different models depending on the used accelerator and the output frame rate. The iGPU (dashed lines) results in higher CPU load than the EdgeTPU (solid lines) for all models.}
  \label{fig:inf_cpu_fps}
\end{figure}
\begin{table}[t]
\caption{Average inference time on the resp. accelerator.} 
\label{tab:model_runtime}
\centering
\begin{threeparttable}
\begin{tabular}{l|c|c|c}
  \toprule %
  Model & Input Resolution & EdgeTPU & iGPU\\
  \midrule %
  RGB segmentation & 849$\,\times\,$481& \SI{40.5}{\milli\second} & \SI{50.0}{\milli\second}\\
  RGB detector & 848$\,\times\,$480 & \SI{17.5}{\milli\second} & \SI{24.0}{\milli\second} \\
  Thermal detector & 640$\,\times\,$512 & \SI{12.0}{\milli\second} & \SI{18.0}{\milli\second} \\
  LiDAR segmentation & 512$\,\times\,$64 & - & \SI{32.0}{\milli\second} \\
  LiDAR segmentation & 1024$\,\times\,$128 & - & \SI{140.0}{\milli\second} \\
  \bottomrule %
\end{tabular}
\end{threeparttable}
\end{table}
\subsection{CNN Model Efficiency}
In real-time systems with limited computational resources, such as UAVs, efficiency is of key importance and resources need to be distributed with care between the different system components. Semantic perception, while important for many high-level tasks, has less severe real-time constraints than, \eg flight control or odometry. It is thus important that the CNN inference uses as few CPU resources as possible to not interfere with the hard real-time constraints of low-level control, localization, and state estimation.
For this, we analyze the CPU load of the employed CNNs for object detection and segmentation, depending on the used accelerator. Although the main computational load of inference is distributed to a dedicated accelerator (EdgeTPU or iGPU), the preparation of input data, data transfer, and post-processing require CPU resources.
This is handled with differing degrees of efficiency\wrt CPU load and depends on the in- and output frame rate, as shown in \reffig{fig:inf_cpu_fps}. Models running on the EdgeTPU produce lower CPU load in all cases while achieving higher or equivalent maximum frame rates. \reftab{tab:model_runtime} shows the average inference latency per model. The LiDAR segmentation is only executed on the iGPU, as the \textit{pixel-shuffle} layer from SalsaNext~\cite{cortinhal_salsanext_2020} is not supported by the EdgeTPU and the model thus cannot be converted to the required 8-bit quantized format.

\begin{table}[t]
\caption{Average CPU load and output frame rate of different model combinations. Image segmentation and detection models run on EdgeTPU and point cloud segmentation on iGPU.} 
\label{tab:multi_inf_cpu_fps}
\centering
\hspace{-3.5em}
\setlength{\extrarowheight}{.1em}
\setlength{\tabcolsep}{0.7em} 
\begin{threeparttable}
\begin{tabular}{ccccccccc}
  \rot{point cloud fusion} & \rot{image fusion} & \rot{point cloud seg.} & \rot{thermal det.} & \rot{RGB det.} & \rot{RGB seg.} & \rot{CPU load} & \rot{avg. FPS (image)} & \rot{avg. FPS (cloud)}\\
  \midrule %
  - & - & - & - & - & \checkmark & \multicolumn{1}{|c|}{\SI{52.5}{\percent}} & \multicolumn{1}{c|}{\SI{21.0}{\hertz}} & - \\
  - & - & - & - & \checkmark & \checkmark & \multicolumn{1}{|c|}{\SI{54.2}{\percent}} & \multicolumn{1}{c|}{\SI{13.2}{\hertz}} & - \\
  - & - & - & \checkmark & \checkmark & \checkmark & \multicolumn{1}{|c|}{\SI{57.3}{\percent}} & \multicolumn{1}{c|}{\SI{12.6}{\hertz}} & - \\
  - & \checkmark & - & \checkmark & \checkmark & \checkmark & \multicolumn{1}{|c|}{\SI{120.6}{\percent}} & \multicolumn{1}{c|}{\SI{9.9}{\hertz}} & - \\
  - & - & \checkmark & \checkmark & \checkmark & \checkmark & \multicolumn{1}{|c|}{\SI{116.6}{\percent}} & \multicolumn{1}{c|}{\SI{12.6}{\hertz}} & \SI{10.0}{\hertz} \\
  - & \checkmark & \checkmark & \checkmark & \checkmark & \checkmark & \multicolumn{1}{|c|}{\SI{180.0}{\percent}} & \multicolumn{1}{c|}{\SI{9.5}{\hertz}} & \SI{10.0}{\hertz} \\
  \checkmark & \checkmark & \checkmark & \checkmark & \checkmark & \checkmark & \multicolumn{1}{|c|}{\SI{204.3}{\percent}} & \multicolumn{1}{c|}{\SI{8.9}{\hertz}} & \SI{9.5}{\hertz} \\
  \bottomrule %
\end{tabular}
\end{threeparttable}
\end{table}
For the following experiments, we choose to run the image CNNs on the EdgeTPU, while the LiDAR segmentation runs on the iGPU at 512$\,\times\,$64 input resolution. \reftab{tab:multi_inf_cpu_fps} shows the average computational load and output rate for different combinations of CNNs. As to be expected, the maximum achievable output frame rate drops and CPU load increases with a growing number of vision models used.
The computation of RGB segmentation and detections, as well as thermal detections, achieves an average frame rate of \SI{12.6}{\hertz} at a CPU load of about \SI{60}{\percent}. The inclusion of the image fusion module almost doubles the CPU utilization while the frame rate drops to \SI{9.9}{\hertz}.
This is due to the transformations and projections necessary to calculate at image resolution for temporal smoothing and to include thermal detection into the fused image segmentation.
The total CPU usage for the fusion of both image and point cloud semantics sums up to about \num{2} CPU cores with an output rate of around \SI{9}{\hertz}. 

Reducing the input frequency to the semantic segmentation and detection can free additional resources for other system components if necessary while still providing semantic image and point cloud, \eg at \SIrange[range-units=single]{1}{5}{\hertz} --- sufficient for many high-level tasks like planning or keyframe-based mapping.

\subsection{Outdoor Experiments}
\label{sec:eval_outdoor}
\begin{figure}[t]
	\centering
	\includegraphics{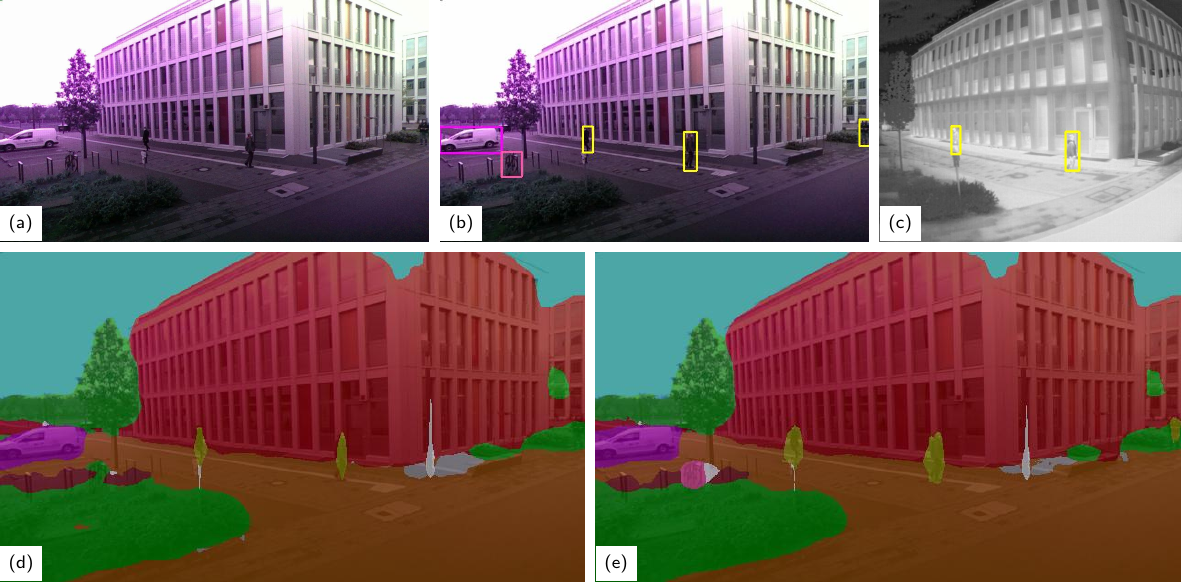}
	\caption{Semantic interpretation of RGB and thermal images: (a) RGB input image, (b) RGB and (c) thermal detections. (d) RGB segmentation. (e) fused segmentation mask. Persons and bicycle, not or only partially segmented in the initial segmentation mask, are fully visible in the fused output.}
	\label{fig:img_semantics}
\end{figure}
In \reffig{fig:img_semantics}, we show results of semantic image fusion for an exemplary scene from our test flights. \reffig{fig:img_semantics} (b) - (d) show the outputs of the individual CNNs. While the large structures are well segmented~(d), the persons are only partially recognized. A bicycle and the person at the right image border are missed altogether. The RGB detector~(b) recognizes all persons and the bicycle. The thermal detector~(c) confirms both person detections inside the thermal camera's FoV. The fused output segmentation mask (e) includes all detections together with the initial segmentation. All persons and the bicycle are clearly visible.

\begin{figure}[t]
	\centering
	\includegraphics[width=\linewidth]{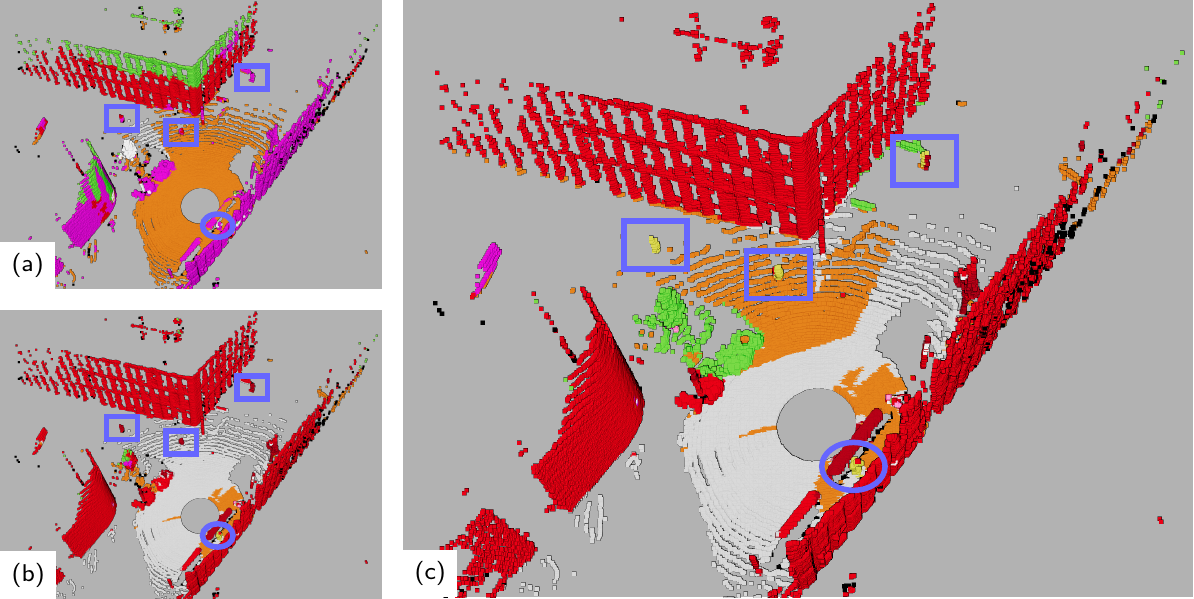}
	\caption{Point cloud segmentation: (a) Initial LiDAR segmentation without adaptation of z and intensity normalization parameters and (b) after adaptation to our dataset mean and std. (c) fused point cloud segmentation. Persons inside (outside) camera FoV are highlighted with blue rectangles (circles).
	After normalization adaptation, the CNN segments large structures well within the LiDAR scan but misses small objects.
	Persons, vegetation, and small structures are well segmented in the fused output scan inside the camera FoV.}
	\label{fig:pointcloud_semantics}
\end{figure}
\reffig{fig:pointcloud_semantics} shows the point cloud segmentation results for the same scene. When using the original LiDAR segmentation, without adaptation of the normalization parameters, parts of buildings are misclassified as vegetation or vehicle. This is likely due to differing vertical field-of-views of our and the trained LiDAR. In the KITTI dataset~\cite{geiger2012cvpr}, the FoV is only \SI{2}{\degree} upwards and $\approx\,$\SI{25}{\degree} downwards (compared to \SI{\pm45}{\degree} of our sensor). Therefore, in SemanticKITTI the top of building structures is rarely visible while treetops are measured from below. Furthermore, the intensity input channel, measured as the magnitude of laser reflection, differs between the sensors, as they use a different wavelength and optical system.
After normalization adaptation, SalsaNext segments the building and road structures well within the LiDAR scan, and the person closest to the sensor is recognized (\reffig{fig:pointcloud_semantics}~(b)). Independent of the normalization, the point cloud network does not detect persons at larger distances, often misclassifying them as barrier or building.
\reffig{fig:pointcloud_semantics}~(c) shows the fused point cloud segmentation, combining image segmentation and detections with the initial point cloud segmentation. Persons, also at larger distances, vegetation, and the car are well segmented in the output scan and exhibit less noise when inside the camera FoV.

\begin{figure}[t]
	\centering
	\includegraphics[width=\linewidth]{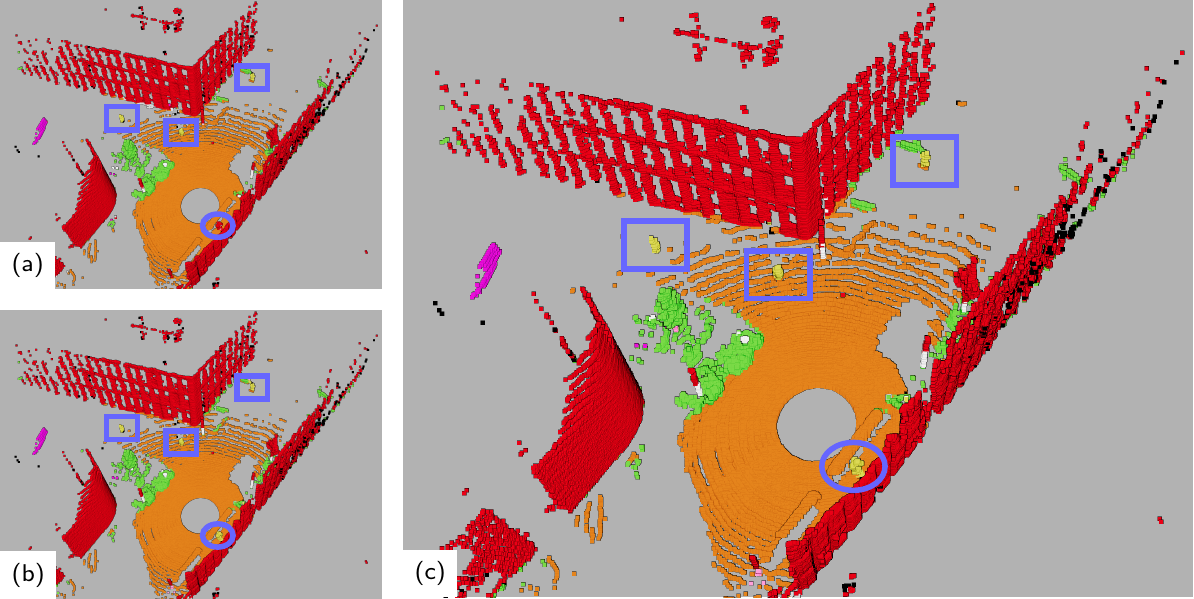}
	\caption{Point cloud segmentation after retraining with label propagation: (a) LiDAR segmentation with intensity input channel and (b) without intensity. (c) fused point cloud segmentation (based on (b)). Persons inside (outside) camera FoV are highlighted with blue rectangles (circles). The scene is accurately segmented in all cases, including persons and small structure. Without intensity input (b) the CNN generalizes better and the bottom person is correctly segmented. Fusing with image semantics gives little gain after retraining.}
	\label{fig:pointcloud_semantics_labelprop}
\end{figure}
The point cloud segmentation after retraining with label propagation is shown in \reffig{fig:pointcloud_semantics_labelprop}.
The scene is segmented very accurately, including persons and small structures, even when using the LiDAR segmentation alone, without fusing the image semantics. The difference between including intensity as input channel or not becomes apparent for the UAV operator (person to the bottom of the scene), who was not annotated in the pseudo-labels used as supervision for label propagation (\cf \refsec{sec:labelprop}). Without intensity input, the generalization works better and this person is also correctly segmented. Additional fusion with the camera semantics makes only little difference after retraining. The LiDAR CNN has learned the relevant segmentation skills from cross-modal supervision.

\begin{figure}[t]
	\centering
	\includegraphics[width=\linewidth]{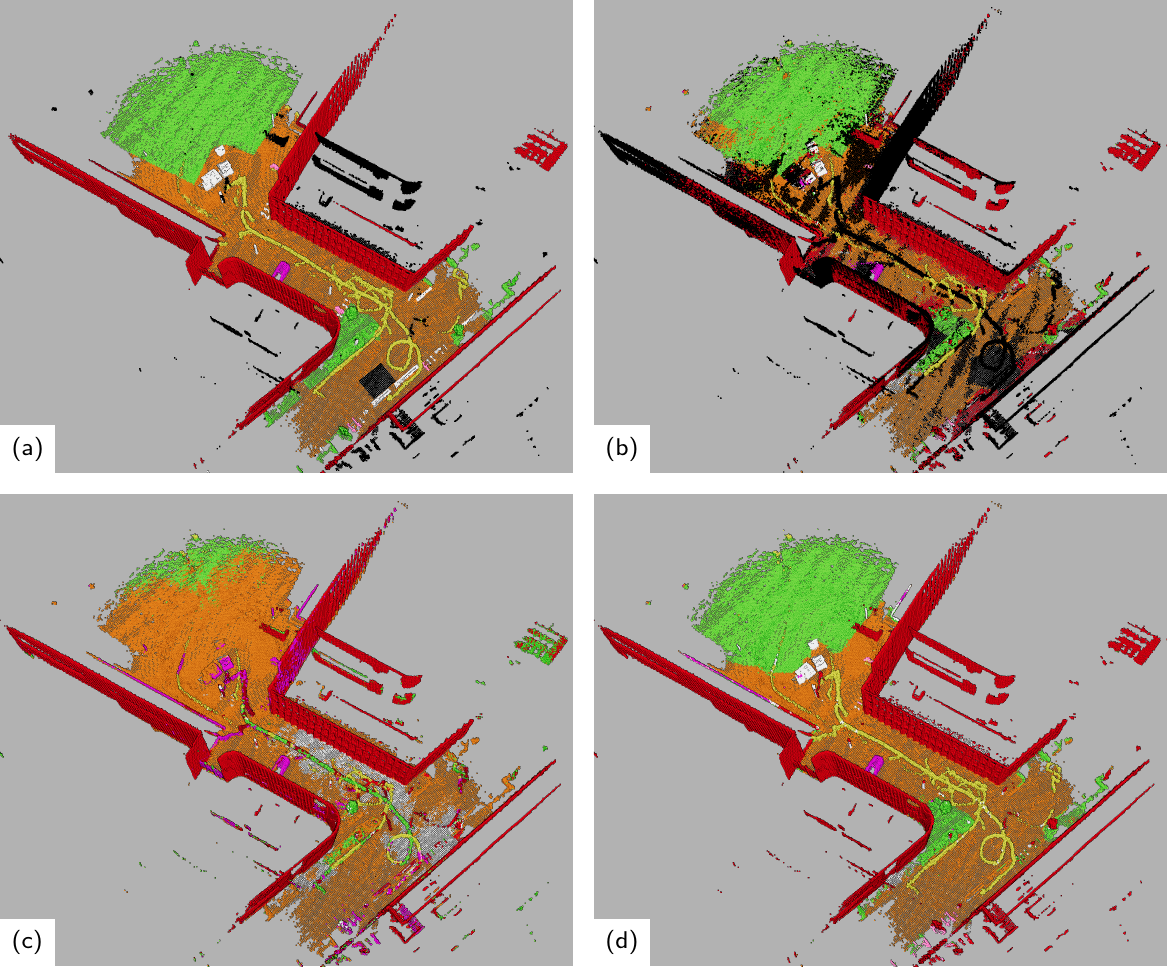}
	\caption{Semantic map of the urban campus outdoor flight. (a) manually annotated ground-truth. (b) map created from scans labeled by projected image semantics only. (c) map created from fused semantic clouds before and (d) after retraining with label propagation. The camera-only map~(b) misses annotations due to the camera's limited FoV but depicts most classes more accurately than (c). The semantic map after retraining (d) depicts all person tracks accurately and is very close to the ground-truth.}
	\label{fig:semantic_map}
\end{figure}
\begin{figure}[t]
	\centering
	\includegraphics{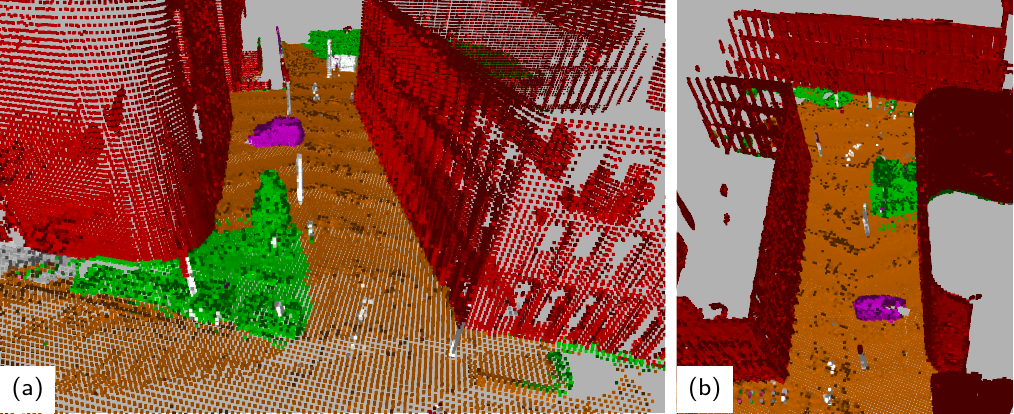}	
	\vspace{-.5em}
	\caption{Closeups from \reffig{fig:semantic_map} for detailed view of the semantic map of the urban campus outdoor flight after removing the tracks of dynamic persons.}
	\label{fig:semantic_map_detail}
\end{figure}
Figure~\ref{fig:semantic_map} depicts the aggregated semantic map of the outdoor test flight with manually annotated semantic labels~(a) and with scans either labeled from image segmentation~(b) or fused semantic point clouds~(c, d). The camera-only map~(b) misses annotations due to the camera's limited FoV but depicts most classes, such as persons, cars, or vegetation, more accurately since the noisier raw point cloud segmentation is not included. The tracks of moving persons are visible in yellow on the maps. Only the track of the operator, who always stayed behind the UAV and thus was not visible in the camera, is not segmented~(b) or mislabeled~(c). The direct segmentation of persons or small structures in the LiDAR scans initially is noisy due to domain adaptation issues with the CNN.
Label propagation for cross-domain supervision is employed to overcome these issues. We use the accurate, but incomplete camera-only map as source for pseudo-labels for retraining the LiDAR segmentation.
The resulting semantic map, \reffig{fig:semantic_map}~(d), aggregated from fused semantic point clouds using the retrained LiDAR CNN, depicts the semantics of the entire scene very accurately with the person tracks completely segmented, including the parts unlabeled in the camera-only map. This underlines the efficiency of label propagation and shows the generalization capabilities of the resulting CNN.
Note, that the manually annotated map is not included during retraining and is only used for evaluation.
For visual assessment of the map quality, \reffig{fig:semantic_map_detail} depicts detailed closeups of the static parts of the final semantic map after removing the dynamic person tracks.

\begin{figure}[t]
	\centering
	\includegraphics[width=0.95\linewidth]{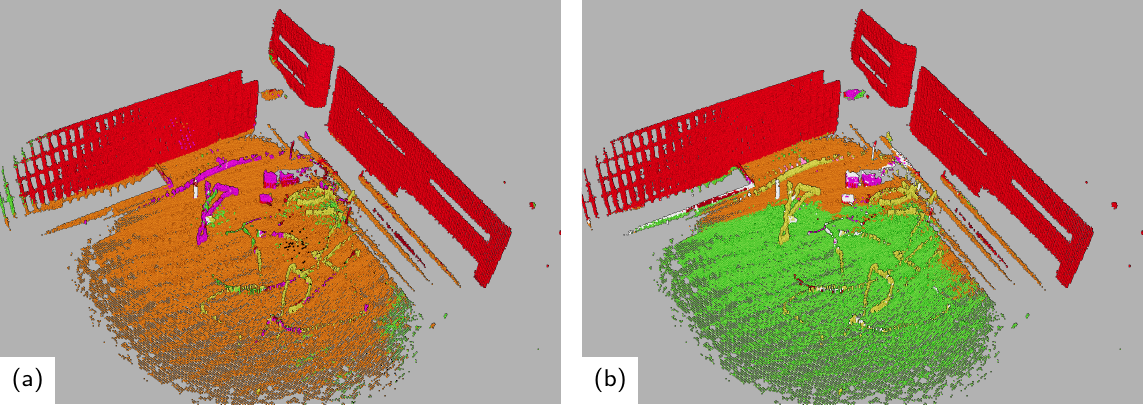}
	\vspace{-.5em}
	\caption{Aggregated maps from fused semantic clouds of another experiment (a) before and (b) after retraining with label propagation. Within the camera frustum (right part of the scene), person detection works sufficiently well, while they are misclassified elsewhere by the original LiDAR CNN. After retraining (b), persons tracks are segmented in the entire scene and the lawn is correctly labeled as vegetation.}
	\label{fig:more_semantic_maps}
\end{figure}
\begin{figure}[t]
	\centering
	\includegraphics[width=0.95\linewidth]{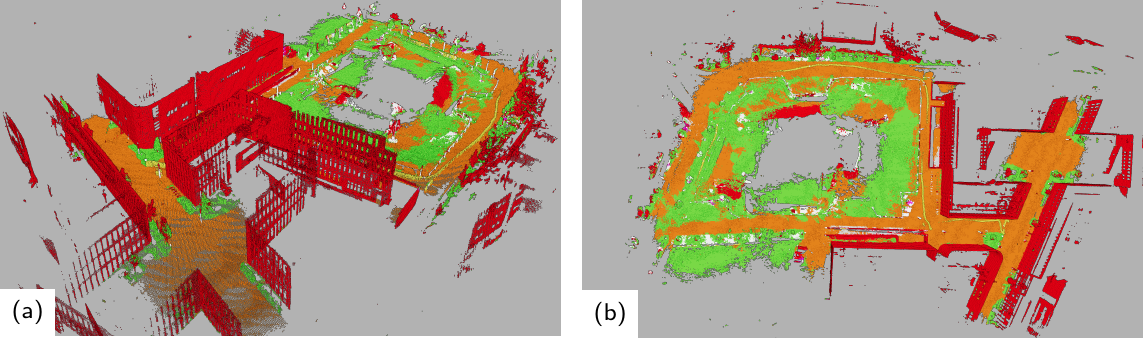}
	\vspace{-.5em}
	\caption{Aggregated semantic map for large area flight: (a) side-view, (b) top-view.}
	\label{fig:baustelle_semantic_map}
\end{figure}
The aggregated maps from fused semantic clouds of two further experiments are shown in \reffig{fig:more_semantic_maps} and \reffig{fig:baustelle_semantic_map}.
For the first flight, the scene semantics are shown (a) before and (b) after retraining with label propagation. Before retraining, person detection works sufficiently well within the camera frustum (right part of the scene), while they are misclassified as vehicle or vegetation elsewhere.
After retraining, person tracks are segmented in all parts of the scene and the lawn is correctly labeled as vegetation.
The semantic map from the second, significantly longer UAV flight around the university campus, using the retrained LiDAR CNN, is shown in \reffig{fig:baustelle_semantic_map}. A coherent 3D semantic representation of the environment can be created also at large scale by our proposed system. Please note, that data from these two flights was not used for label propagation.

\subsubsection*{IoU Evaluation}
To quantitatively evaluate the coherence of different point cloud segmentations, we calculate the intersection-over-union (IoU):
\begin{align}
IoU_c &= \frac{TP_c}{TP_c+FP_c+FN_c}\,,
\end{align}
where TP, FP, and FN are the true positives, false positives, and false negatives, respectively.
We compare for each segmented point its $\argmax$ class against the corresponding aggregated voxel label and average per class over the whole dataset. \reftab{tab:iou} shows the results for all classes that occur for a significant number of points in our recorded data. We use the manually annotated aggregated semantic map with a voxel size of \SI{25}{\centi\metre} as ground-truth (cf. \reffig{fig:semantic_map}~(a)).
\begin{table}[t]
\caption{Average IoU per class (in \si{\percent}) and mean IoU for different point cloud segmentations, LiDAR CNN input resolutions, and FoV measured against manually annotated semantic map.} 
\label{tab:iou}
\centering
\setlength{\tabcolsep}{0.1em} 
\begin{threeparttable}
\begin{tabular}{l|c|c|cccccccc}
  \toprule
  Method & Res & FoV & Build. & Road & Veg. & Pers. & Bike & Car & Obj. & Mean\\
  \midrule %
  single w/o adapt$^*$& 512 & LiDAR & 32.7 & 72.9 & 2.2 & 2.7 & 3.7 & 0.6 & 3.4 & 16.9\\
  single w/ adapt$^*$& 512 & LiDAR & 83.4 & 75.5 & 2.2 & 17.7 & 7.2 & 2.8 & 5.5 & 27.8\\
  single w/ adapt & 1024\,&\,LiDAR\,& 82.7 & 71.8 & 4.0 & 43.3 & 6.3 & 4.1 & 7.4 & 31.4\\
  fused & 512 & LiDAR & \textbf{84.3} & \textbf{77.0} & 17.8 & 23.3 & \textbf{8.4} & 3.9 & 8.0 & 31.8\\
  fused & 1024\,& LiDAR & 84.0 & 73.8 & \textbf{18.7} & \textbf{47.2} & 7.0 & \textbf{5.3} & \textbf{9.5} & \textbf{35.1}\\
  \midrule
  fused & 512 & camera & \textit{94.0} & \textit{62.2} & \textit{48.6} & \textit{36.6} & \textit{17.1} & \textit{34.0} & \textit{32.1} & \textit{46.4}\\
  fused & 1024\,& camera & \textit{\textbf{94.3}} & \textit{63.8} & \textit{50.3} & \textit{36.9} & \textit{\textbf{17.6}} & \textit{35.2} & \textit{37.1} & \textit{47.9}\\
  img proj. & n/a & camera & \textit{92.6} & \textit{\textbf{76.5}} & \textit{\textbf{77.1}} & \textit{\textbf{37.1}} & \textit{17.2} & \textit{\textbf{46.4}} & \textit{\textbf{39.1}} & \textit{\textbf{55.1}}\\
  \bottomrule %
\end{tabular}
\end{threeparttable}
\\\vspace{0.5em}
{\linespread{-0.15}\footnotesize\raggedright $^*$differences\wrt original results from~\cite{bultmann2021real} are due to a bug-fix in exporting the \textit{pixel-shuffle} layer of the LiDAR CNN to the employed OpenVINO inference framework.\par}
\end{table}

Applying the proposed adaptation of the normalization parameters improves the overall segmentation accuracy. The improvement is most significant for the building class, as the top half of the buildings are correctly labeled (\cf \reffig{fig:pointcloud_semantics}).
The fused semantic cloud improves the segmentation coherence for all classes, especially for persons and vegetation. Persons and small objects, such as bicycles, are more reliably detected in the RGB and thermal camera modalities thus improving the fused output inside the camera frustum.
Using a higher input resolution for LiDAR segmentation gives significant improvements for the person class (\SI{43.3}{\percent} vs. \SI{17.7}{\percent}) and a small improvement of mean IoU.
The track of the operator, who always stayed behind the UAV and was not visible in the camera, is often misclassified, significantly impacting the mIoU of the person class as it constitutes a large number of the points annotated as person in the ground-truth map. The LiDAR segmentation with full input resolution segments larger parts of the operator track correctly.
Results for the semantic cloud from the projected image segmentation and the fused semantic cloud evaluated for the reduced FoV of the camera show significantly improved mIoU values also for vegetation, cars, and other foreground objects.

The semantic fusion initially proposed in~\cite{bultmann2021real} successfully creates a coherently labeled 3D semantic interpretation of global structure in the full \SI{360}{\degree} LiDAR FoV and for both global structure and small dynamic objects in the camera FoV.

\begin{table}[t]
\caption{Label Propagation: Average IoU per class (in \si{\percent}) and mean IoU for point cloud segmentation with label propagation measured against the manually annotated semantic map.} 
\label{tab:iou_labelprop}
\centering
\setlength{\tabcolsep}{0.25em} 
\begin{threeparttable}
\begin{tabular}{l|c|c|cccccccc}
  \toprule
  Method & Res & Intensity & Build. & Road & Veg. & Pers. & Bike & Car & Obj. & Mean\\
  \midrule %
  single & 512 & \checkmark & 95.8 & \textbf{88.7} & 77.6 & 49.4 & 14.4 & 56.2 & 22.8 & 57.8\\
  single & 512 & - & 95.5 & 88.3 & 75.3 & 65.0 & 19.5 & \textbf{66.9} & \textbf{31.6} & \textbf{63.2}\\
  single & 1024 & - & 95.4 & 88.5 & 74.1 & 67.7 & \textbf{21.3} & 55.2 & 22.9 & 60.7\\
  fused  & 512 & \checkmark & \textbf{95.9} & \textbf{88.7} & \textbf{77.8} & 49.5 & 14.5 & 55.2 & 22.7 & 57.8\\
  fused  & 512 & - & 95.6 & 88.4 & 75.8 & 65.4 & 19.1 & 60.8 & 31.2 & 62.3 \\
  fused  & 1024 & - & 95.5 & 88.5 & 74.4 & \textbf{67.8} & 21.2 & 53.7 & 22.7 & 60.5\\
  \bottomrule %
\end{tabular}
\end{threeparttable}
\end{table}
To improve the accuracy for the difficult semantic classes in the entire FoV, label propagation is used for retraining the LiDAR segmentation with pseudo-labels obtained from the camera-only map (\cf \refsec{sec:labelprop}). The results are shown in \reftab{tab:iou_labelprop}.
The segmentation accuracy for the whole \SI{360}{\degree} horizontal FoV almost doubles from \SI{31.8}{\percent} to \SI{62.3}{\percent} and also is significantly higher than the \SI{55.1}{\percent} previously achieved in the camera FoV only.
The retrained LiDAR segmentation CNN generalizes better without using the intensity input channel for the person and small object classes. Fusing the LiDAR segmentation with the image semantics gives a small gain for the person class but overall performs slightly worse.
This underlines that the LiDAR CNN has learned the relevant segmentation skill from the RGB and thermal image modalities through cross-modal supervision during retraining.
Small gains from an increased input resolution can still be observed for person and bicycle classes, but the mean IoU does not improve.
Furthermore, the inference speed drops below the \SI{10}{\hertz} measurement frequency of the LiDAR at the 1024$\,\times\,$128 input resolution (\cf \reftab{tab:model_runtime}). Therefore, we use the CNN at 512$\,\times\,$64 input resolution in our online experiments.

\begin{table}[t]
\caption{Ablation on data used for label propagation (LP): Mean IoU (in \si{\percent}) of point cloud segmentation using different datasets measured against the manually annotated map.} 
\label{tab:iou_data}
\centering
\linespread{1.25}\selectfont
\setlength{\tabcolsep}{0.5em} 
\begin{threeparttable}
\begin{tabular}{l|c|c|c}
  \toprule
  Datasets & LP & Intensity & Mean IoU\\
  \midrule %
  Sem.KITTI (pretrained~\cite{cortinhal_salsanext_2020}) & - & \checkmark & 27.8 \\
  Sem.KITTI + Paris & - & \checkmark & 32.7 \\
  Sem.KITTI + Paris & - & - & 41.2 \\
  Sem.KITTI + Paris & fused map & \checkmark & 37.5 \\
  Sem.KITTI + Paris & fused map & - & 38.5 \\
  Sem.KITTI + Paris & single scan & \checkmark & 49.4 \\
  Sem.KITTI + Paris & single scan & - & 57.0 \\
  Sem.KITTI & camonly map & \checkmark & 57.6 \\
  Sem.KITTI & camonly map & - & 60.4 \\
  Sem.KITTI + Paris & camonly map & \checkmark & 57.8 \\
  Sem.KITTI + Paris & camonly map & - & \textbf{63.2}\\
  \bottomrule %
\end{tabular}
\end{threeparttable}
\end{table}
We further perform an ablation study on the employed training data and pseudo-label sources in \reftab{tab:iou_data}.
The addition of data from Paris-CARLA-3D~\cite{deschaud2021paris}, which covers the entire sensor FoV, improves the mean IoU by \SIrange{5}{13}{\percent}, depending on whether the intensity channel is used.
As observed in the previous analysis, networks not using the intensity input channel generalize better to our dataset.
The most significant gain is achieved by adding data recorded from our sensor, using label propagation from the camera modality as cross-domain supervision. Furthermore, this reduces the noticeable differences for the intensity channel.
Pseudo-labels obtained from the fused semantic map, including the noisier raw point cloud segmentation, are too imprecise to achieve significant improvements after retraining.
Using pseudo-labels from the camera overlay of single scans improves the segmentation quality despite only a small portion of each scan being labeled (\cf \reffig{fig:labelprop}~(b)). The best results are achieved using pseudo-labels from the temporally and spatially aggregated camera-only semantic map, with an improvement of \SI{35.4}{\percent} over the initial point cloud segmentation in our scenario.
Using only own data for retraining, without Paris-CARLA-3D, performs slightly worse, as the training data is less diverse.

\begin{table}[t]
\caption{Mean IoU (in \si{\percent}) of aggregated semantic maps measured against the manually annotated ground-truth.} 
\label{tab:iou_map}
\centering
\linespread{1.2}\selectfont
\begin{threeparttable}
\begin{tabular}{l|m{1.75cm}|m{1.9cm}|m{2.2cm}|m{2.9cm}}
  \toprule
  Method & \centering fused cloud & \centering camera-only & \centering camera-only (camera FoV) & \centering fused cloud w/ LP\tabularnewline
  \midrule %
  mIoU   & \centering 29.5 & \centering 41.8 & \centering 60.3 & \centering 68.0 \tabularnewline
  \bottomrule %
\end{tabular}
\end{threeparttable}
\end{table}
While in the previous evaluations, IoU was calculated for individual semantic point clouds and averaged over the whole dataset, we show the IoU results of the aggregated semantic maps in \reftab{tab:iou_map}. As for the single scans, the semantic segmentation of the camera modalities is significantly more accurate than the fused cloud including the raw point cloud segmentation. For the \textit{camera FoV} evaluation, we use only points inside the camera frustum and do not count unlabeled points as false negatives. With label propagation, the improvement upon the camera-only map, used as source for pseudo-labels during retraining, amounts to $\approx\SI{8}{\percent}$ for the limited camera FoV and $\approx\SI{26}{\percent}$ for the full FoV.
The accurate automatically generated supervision from the image domain with narrow FoV could be generalized to the full LiDAR FoV via label propagation.

\subsubsection*{Disaster Test Site}
For qualitative evaluation, further flights were conducted on a disaster test site of the German Rescue Robotics Center~\cite{drz2021ssrr}.
These experiments show the generalization capabilities of the proposed system to environments significantly different from the urban campus area, where the data used for label propagation was recorded. The retrained LiDAR CNN is directly employed for these experiments, without any further adaptation.

\begin{figure}[t]
	\centering
	\includegraphics[width=\linewidth]{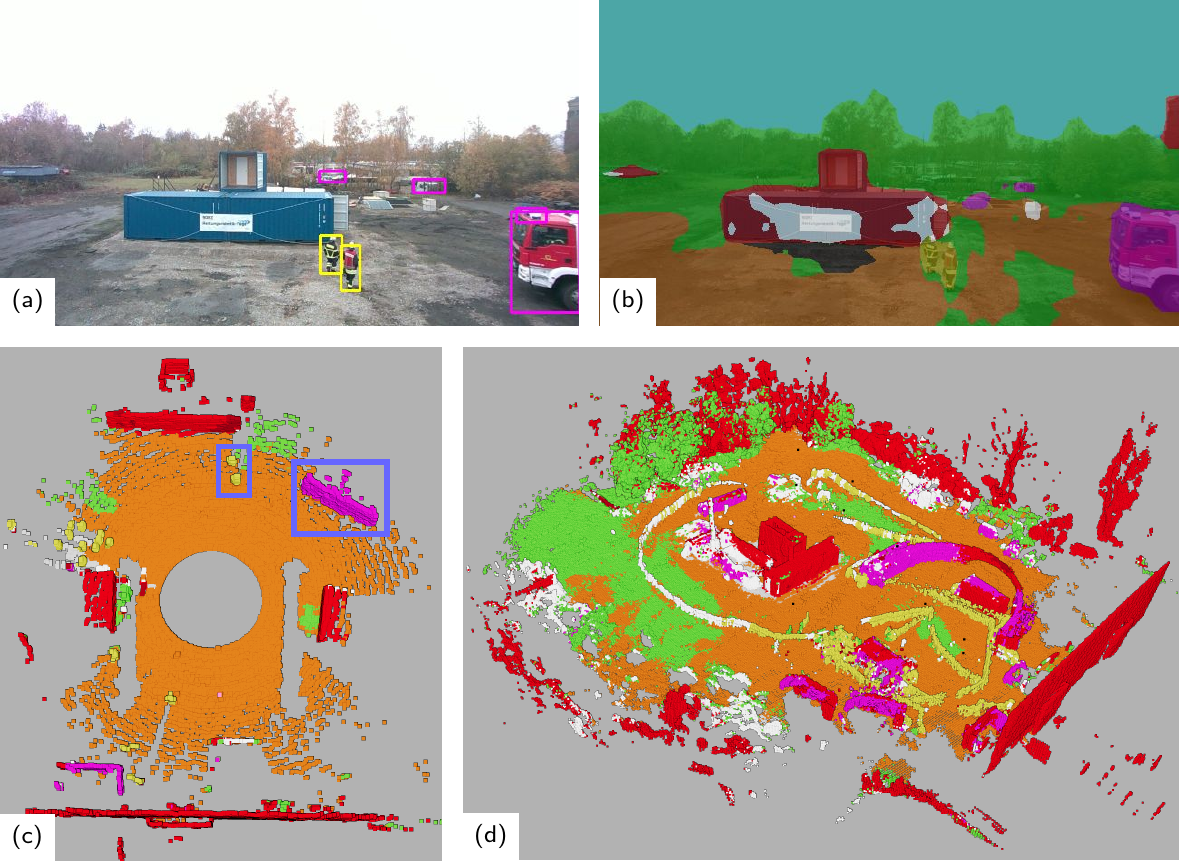}
	\vspace{-2.em}
	\caption{Semantic perception on disaster test site: (a) RGB detections, (b) fused RGB segmentation, (c) fused semantic cloud, (d) aggregated semantic cloud. Persons and fire truck inside the camera FoV are highlighted with blue rectangles in (c).}
	\label{fig:drz_demo_eval}
	\vspace{-.5em}
\end{figure}
Figure~\ref{fig:drz_demo_eval} shows the semantic perception of the disaster test site. Multiple persons, a fire truck, and also further away cars are reliably detected in the image and included into the fused image segmentation and point cloud. The point cloud segmentation also labels persons not visible in the camera correctly. Figure~\ref{fig:drz_demo_eval}~(d) depicts the aggregated semantic map of the scenario. Some noise is visible in one of the person tracks and the higher trees are erroneously labeled as building structure, but the overall perception remains coherent also in this challenging scenario.

\begin{figure}[t]
	\centering
	\includegraphics[width=\linewidth]{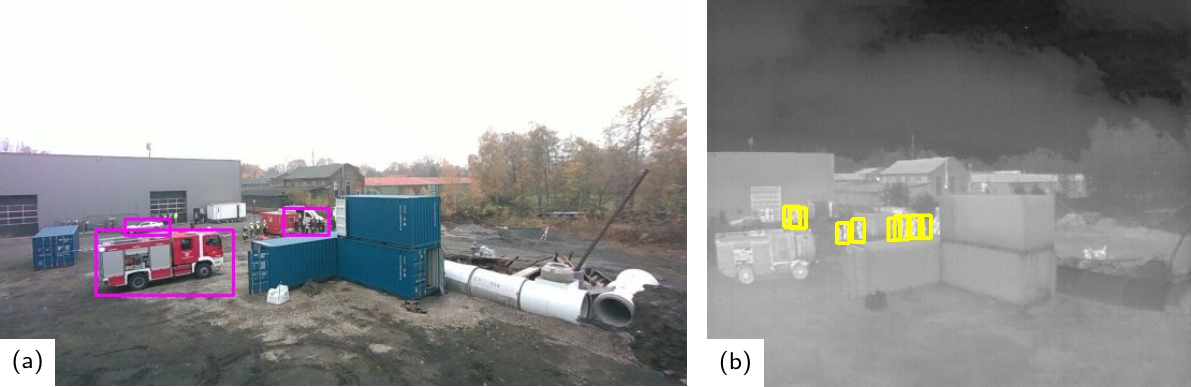}
	\caption{Comparison of (a) RGB and (b) thermal detections on disaster test site. Persons are detected in the thermal image also at very high distance, while only the larger vehicles are detected in the RGB image.}
	\vspace{-1.em}
	\label{fig:drz_dets}
	\vspace{-.3em}
\end{figure}
Figure~\ref{fig:drz_dets} highlights the benefits of the thermal camera also at daylight: Persons are detected at high distances, while in the RGB image only the larger vehicles are detected. For the thermal camera, transfer from the training dataset with autonomous driving scenarios to the aerial perspectives of the UAV flights was possible without explicit domain transfer techniques, as the sensor characteristics in the dataset and of the employed thermal camera are similar.

To further improve the results for the disaster test site, another iteration of label propagation could be performed, using pseudo-labels automatically obtained from the aggregated semantic map of the environments.
 
\section{Conclusion}
\label{sec:Conclusion}
In this work, we presented a UAV system for semantic image and point cloud analysis as well as multi-modal semantic fusion.
The inference of the lightweight CNN models runs onboard the UAV computer, employing an inference accelerator and the integrated GPU of the main processor for computation. 
The EdgeTPU performs inference in 8-bit quantized mode and showed more efficient CPU usage. The iGPU is more flexible, \eg to directly run pre-trained models, as it uses 16- or 32-bit floating-point precision and does not require model quantization.

The proposed framework for semantic scene analysis provides a 2D image segmentation overlay and a 3D semantically labeled point cloud which is further aggregated into an allocentric semantic map.
The initial point cloud segmentation suffered from domain adaptation issues since available large-scale training datasets stem from autonomous driving scenarios with different viewpoints and sensors more focused towards the ground compared to the LiDAR sensor of the UAV.
We addressed this issue by retraining the LiDAR segmentation CNN with data captured on our UAV using pseudo-labels automatically obtained from the aggregated semantic map. The pseudo-labels thereby stem from the RGB and thermal camera modalities, providing cross-domain supervision for the 3D point cloud.
With label propagation, the 3D segmentation accuracy of the proposed system significantly improves for the full LiDAR FoV.
We evaluated the system in real-world experiments in an urban environment and at a disaster test site, showing coherent semantic perception of diverse and challenging scenes.

Future work includes applying label propagation to retrain also the image CNNs to achieve better tasks-specific performance and adding RGB color channels to the input of the point cloud segmentation \eg using a \SI{360}{\degree} fisheye camera together with the LiDAR sensor. 
\section*{Acknowledgments}
This work has been supported by the German Federal Ministry of Education and Research (BMBF) in the project ``Kompetenzzentrum: Aufbau des Deutschen Rettungsrobotik-Zentrums (A-DRZ)'', grant No. 13N14859.

\bibliography{literature}

\begin{thebibliography}{10}
\expandafter\ifx\csname url\endcsname\relax
  \def\url#1{\texttt{#1}}\fi
\expandafter\ifx\csname urlprefix\endcsname\relax\def\urlprefix{URL }\fi
\expandafter\ifx\csname href\endcsname\relax
  \def\href#1#2{#2} \def\path#1{#1}\fi

\bibitem{drz2021ssrr}
I.~Kruijff-Korbayová, R.~Grafe, N.~Heidemann, A.~Berrang, C.~Hussung,
  C.~Willms, P.~Fettke, M.~Beul, J.~Quenzel, D.~Schleich, S.~Behnke,
  J.~Tiemann, J.~Güldenring, M.~Patchou, C.~Arendt, C.~Wietfeld, K.~Daun,
  M.~Schnaubelt, O.~von Stryk, A.~Lel, A.~Miller, C.~Röhrig, T.~Straßmann,
  T.~Barz, S.~Soltau, F.~Kremer, S.~Rilling, R.~Haseloff, S.~Grobelny,
  A.~Leinweber, G.~Senkowski, M.~Thurow, D.~Slomma, H.~Surmann, German rescue
  robotics center ({DRZ}): A holistic approach for robotic systems assisting in
  emergency response, in: IEEE Int. Symposium on Safety, Security and Rescue
  Robotics (SSRR), 2021, pp. 138--145.
\newblock \href {http://dx.doi.org/10.1109/SSRR53300.2021.9597869}
  {\path{doi:10.1109/SSRR53300.2021.9597869}}.

\bibitem{nguyen_mavnet_2019}
T.~Nguyen, S.~S. Shivakumar, I.~D. Miller, J.~Keller, E.~S. Lee, A.~Zhou,
  T.~Özaslan, G.~Loianno, J.~H. Harwood, J.~Wozencraft, C.~J. Taylor,
  V.~Kumar, {MAVNet}: An effective semantic segmentation micro-network for
  {MAV}-based tasks, IEEE Robotics and Automation Letters (RA-L) 4~(4) (2019)
  3908--3915.
\newblock \href {http://dx.doi.org/10.1109/LRA.2019.2928734}
  {\path{doi:10.1109/LRA.2019.2928734}}.

\bibitem{bartolomei_perception-aware_2020}
L.~Bartolomei, L.~Teixeira, M.~Chli, Perception-aware path planning for {UAVs}
  using semantic segmentation, in: IEEE/RSJ Int. Conf. on Intelligent Robots
  and Systems (IROS), 2020, pp. 5808--5815.
\newblock \href {http://dx.doi.org/10.1109/IROS45743.2020.9341347}
  {\path{doi:10.1109/IROS45743.2020.9341347}}.

\bibitem{chen_suma_2019}
X.~Chen, A.~Milioto, E.~Palazzolo, P.~Giguere, J.~Behley, C.~Stachniss,
  {SuMa}++: Efficient {LiDAR}-based semantic {SLAM}, in: IEEE/RSJ Int. Conf. on
  Intelligent Robots and Systems (IROS), 2019, pp. 4530--4537.
\newblock \href {http://dx.doi.org/10.1109/IROS40897.2019.8967704}
  {\path{doi:10.1109/IROS40897.2019.8967704}}.

\bibitem{bultmann2021real}
S.~Bultmann, J.~Quenzel, S.~Behnke, Real-time multi-modal semantic fusion on
  unmanned aerial vehicles, in: Europ. Conf. on Mobile Robots (ECMR), 2021, pp.
  1--8.
\newblock \href {http://dx.doi.org/10.1109/ECMR50962.2021.9568812}
  {\path{doi:10.1109/ECMR50962.2021.9568812}}.

\bibitem{mobilenetv2_2018}
M.~Sandler, A.~Howard, M.~Zhu, A.~Zhmoginov, L.-C. Chen, {MobileNetV2}:
  Inverted residuals and linear bottlenecks, in: IEEE Conf. on Computer Vision
  and Pattern Recognition (CVPR), 2018, pp. 4510--4520.
\newblock \href {http://dx.doi.org/10.1109/CVPR.2018.00474}
  {\path{doi:10.1109/CVPR.2018.00474}}.

\bibitem{mobilenetv32019}
A.~Howard, M.~Sandler, G.~Chu, L.-C. Chen, B.~Chen, M.~Tan, W.~Wang, Y.~Zhu,
  R.~Pang, V.~Vasudevan, Q.~V. Le, H.~Adam, {Searching for MobileNetV3}, in:
  IEEE Int. Conf. on Computer Vision (ICCV), 2019, pp. 1314--1324.
\newblock \href {http://dx.doi.org/10.1109/ICCV.2019.00140}
  {\path{doi:10.1109/ICCV.2019.00140}}.

\bibitem{he_deep_2016}
K.~He, X.~Zhang, S.~Ren, J.~Sun, {Deep residual learning for image
  recognition}, in: IEEE Conf. on Computer Vision and Pattern Recognition
  (CVPR), 2016, pp. 770--778.
\newblock \href {http://dx.doi.org/10.1109/CVPR.2016.90}
  {\path{doi:10.1109/CVPR.2016.90}}.

\bibitem{liu_ssd_2016}
W.~Liu, D.~Anguelov, D.~Erhan, C.~Szegedy, S.~Reed, C.-Y. Fu, A.~C. Berg, {SSD:
  Single shot multibox detector}, in: Europ. Conf. on Computer Vision (ECCV),
  2016, pp. 21--37.
\newblock \href {http://dx.doi.org/10.1007/978-3-319-46448-0_2}
  {\path{doi:10.1007/978-3-319-46448-0_2}}.

\bibitem{Redmon_YOLO_2016}
J.~Redmon, S.~Divvala, R.~Girshick, A.~Farhadi, You only look once: Unified,
  real-time object detection, in: IEEE Conf. on Computer Vision and Pattern
  Recognition (CVPR), 2016, pp. 779--788.
\newblock \href {http://dx.doi.org/10.1109/CVPR.2016.91}
  {\path{doi:10.1109/CVPR.2016.91}}.

\bibitem{Zhang_2019_ICCV}
P.~Zhang, Y.~Zhong, X.~Li, {SlimYOLOv3}: Narrower, faster and better for
  real-time {UAV} applications, in: IEEE Int. Conf. on Computer Vision
  Workshops (ICCVW), 2019, pp. 37--45.
\newblock \href {http://dx.doi.org/10.1109/ICCVW.2019.00011}
  {\path{doi:10.1109/ICCVW.2019.00011}}.

\bibitem{xiong_mobiledets_2021}
Y.~Xiong, H.~Liu, S.~Gupta, B.~Akin, G.~Bender, Y.~Wang, P.-J. Kindermans,
  M.~Tan, V.~Singh, B.~Chen, Mobiledets: Searching for object detection
  architectures for mobile accelerators, in: IEEE Conf. on Computer Vision and
  Pattern Recognition (CVPR), 2021, pp. 3825--3834.
\newblock \href {http://dx.doi.org/10.1109/CVPR46437.2021.00382}
  {\path{doi:10.1109/CVPR46437.2021.00382}}.

\bibitem{weedNet2018}
I.~Sa, Z.~Chen, M.~Popović, R.~Khanna, F.~Liebisch, J.~Nieto, R.~Siegwart,
  {weedNet}: Dense semantic weed classification using multispectral images and
  {MAV} for smart farming, IEEE Robotics and Automation Letters (RA-L) 3~(1)
  (2018) 588--595.
\newblock \href {http://dx.doi.org/10.1109/LRA.2017.2774979}
  {\path{doi:10.1109/LRA.2017.2774979}}.

\bibitem{deeplabv3plus2018}
L.-C. Chen, Y.~Zhu, G.~Papandreou, F.~Schroff, H.~Adam, Encoder-decoder with
  atrous separable convolution for semantic image segmentation, in: Europ.
  Conf. on Computer Vision (ECCV), 2018, pp. 833--841.
\newblock \href {http://dx.doi.org/10.1007/978-3-030-01234-2_49}
  {\path{doi:10.1007/978-3-030-01234-2_49}}.

\bibitem{cortinhal_salsanext_2020}
T.~Cortinhal, G.~Tzelepis, E.~E. Aksoy, {SalsaNext}: Fast, uncertainty-aware
  semantic segmentation of {LiDAR} point clouds, in: Int. Symposium on Visual
  Computing, 2020, pp. 207--222.
\newblock \href {http://dx.doi.org/10.1007/978-3-030-64559-5_16}
  {\path{doi:10.1007/978-3-030-64559-5_16}}.

\bibitem{milioto_rangenet_2019}
A.~Milioto, I.~Vizzo, J.~Behley, C.~Stachniss, {RangeNet} ++: Fast and accurate
  {LiDAR} semantic segmentation, in: IEEE/RSJ Int. Conf. on Intelligent Robots
  and Systems (IROS), 2019, pp. 4213--4220.
\newblock \href {http://dx.doi.org/10.1109/IROS40897.2019.8967762}
  {\path{doi:10.1109/IROS40897.2019.8967762}}.

\bibitem{xu_squeezessegv3_2020}
C.~Xu, B.~Wu, Z.~Wang, W.~Zhan, P.~Vajda, K.~Keutzer, M.~Tomizuka,
  {SqueezeSegV3}: Spatially-adaptive convolution for efficient point-cloud
  segmentation, in: Europ. Conf. on Computer Vision (ECCV), 2020, pp. 1--19.
\newblock \href {http://dx.doi.org/10.1007/978-3-030-58604-1_1}
  {\path{doi:10.1007/978-3-030-58604-1_1}}.

\bibitem{qi2021offboard}
C.~R. Qi, Y.~Zhou, M.~Najibi, P.~Sun, K.~Vo, B.~Deng, D.~Anguelov, Offboard
  {3D} object detection from point cloud sequences, in: IEEE Conf. on Computer
  Vision and Pattern Recognition (CVPR), 2021, pp. 6134--6144.
\newblock \href {http://dx.doi.org/10.1109/CVPR46437.2021.00607}
  {\path{doi:10.1109/CVPR46437.2021.00607}}.

\bibitem{behley2019iccv}
J.~Behley, M.~Garbade, A.~Milioto, J.~Quenzel, S.~Behnke, C.~Stachniss,
  J.~Gall, {SemanticKITTI}: A dataset for semantic scene understanding of
  {LiDAR} sequences, in: IEEE Int. Conf. on Computer Vision (ICCV), 2019, pp.
  9296--9306.
\newblock \href {http://dx.doi.org/10.1109/ICCV.2019.00939}
  {\path{doi:10.1109/ICCV.2019.00939}}.

\bibitem{Xu_pointfusion_2018}
D.~Xu, D.~Anguelov, A.~Jain, {PointFusion}: Deep sensor fusion for {3D}
  bounding box estimation, in: IEEE Conf. on Computer Vision and Pattern
  Recognition (CVPR), 2018, pp. 244--253.
\newblock \href {http://dx.doi.org/10.1109/CVPR.2018.00033}
  {\path{doi:10.1109/CVPR.2018.00033}}.

\bibitem{Qi_2017_pointnet}
C.~R. Qi, H.~Su, K.~Mo, L.~J. Guibas, {PointNet}: Deep learning on point sets
  for {3D} classification and segmentation, in: IEEE Conf. on Computer Vision
  and Pattern Recognition (CVPR), 2017, pp. 77--85.
\newblock \href {http://dx.doi.org/10.1109/CVPR.2017.16}
  {\path{doi:10.1109/CVPR.2017.16}}.

\bibitem{Meyer_2019_CVPR_Workshops}
G.~P. Meyer, J.~Charland, D.~Hegde, A.~Laddha, C.~Vallespi-Gonzalez, Sensor
  fusion for joint {3D} object detection and semantic segmentation, in: IEEE
  Conf. on Computer Vision and Pattern Recognition (CVPR) Workshops, 2019, pp.
  1230--1237.
\newblock \href {http://dx.doi.org/10.1109/CVPRW.2019.00162}
  {\path{doi:10.1109/CVPRW.2019.00162}}.

\bibitem{Vora_2020_CVPR}
S.~Vora, A.~H. Lang, B.~Helou, O.~Beijbom, {PointPainting}: Sequential fusion
  for {3D} object detection, in: IEEE Conf. on Computer Vision and Pattern
  Recognition (CVPR), 2020, pp. 4603--4611.
\newblock \href {http://dx.doi.org/10.1109/CVPR42600.2020.00466}
  {\path{doi:10.1109/CVPR42600.2020.00466}}.

\bibitem{zhao2021lifseg}
L.~Zhao, H.~Zhou, X.~Zhu, X.~Song, H.~Li, W.~Tao, {LIF-Seg}: Lidar and camera
  image fusion for {3D} {LiDAR} semantic segmentation, preprint
  arXiv:2108.07511.

\bibitem{zhu2020cylindrical}
X.~Zhu, H.~Zhou, T.~Wang, F.~Hong, Y.~Ma, W.~Li, H.~Li, D.~Lin, Cylindrical and
  asymmetrical {3D} convolution networks for {LiDAR} segmentation, in: IEEE
  Conf. on Computer Vision and Pattern Recognition (CVPR), 2021, pp.
  9939--9948.
\newblock \href {http://dx.doi.org/10.1109/CVPR46437.2021.00981}
  {\path{doi:10.1109/CVPR46437.2021.00981}}.

\bibitem{mccormac_semanticfusion_2017}
J.~McCormac, A.~Handa, A.~Davison, S.~Leutenegger, {SemanticFusion}: Dense {3D}
  semantic mapping with convolutional neural networks, in: IEEE Int. Conf. on
  Robotics and Automation (ICRA), 2017, pp. 4628--4635.
\newblock \href {http://dx.doi.org/10.1109/ICRA.2017.7989538}
  {\path{doi:10.1109/ICRA.2017.7989538}}.

\bibitem{whelan2015elasticfusion}
T.~Whelan, S.~Leutenegger, R.~Salas-Moreno, B.~Glocker, A.~Davison,
  {ElasticFusion}: Dense {SLAM} without a pose graph, in: Robotics: Science and
  Systems (RSS), 2015.
\newblock \href {http://dx.doi.org/10.15607/RSS.2015.XI.001}
  {\path{doi:10.15607/RSS.2015.XI.001}}.

\bibitem{rosinol2020kimera}
A.~Rosinol, M.~Abate, Y.~Chang, L.~Carlone, {Kimera}: An open-source library
  for real-time metric-semantic localization and mapping, in: IEEE Int. Conf.
  on Robotics and Automation (ICRA), 2020, pp. 1689--1696.
\newblock \href {http://dx.doi.org/10.1109/ICRA40945.2020.9196885}
  {\path{doi:10.1109/ICRA40945.2020.9196885}}.

\bibitem{oleynikova2017voxblox}
H.~Oleynikova, Z.~Taylor, M.~Fehr, R.~Siegwart, J.~Nieto, {Voxblox}:
  Incremental {3D} euclidean signed distance fields for on-board {MAV}
  planning, in: IEEE/RSJ Int. Conf. on Intelligent Robots and Systems (IROS),
  2017.

\bibitem{grinvald2019vol}
M.~Grinvald, F.~Furrer, T.~Novkovic, J.~J. Chung, C.~Cadena, R.~Siegwart,
  J.~Nieto, Volumetric instance-aware semantic mapping and {3D} object
  discovery, IEEE Robotics and Automation Letters (RA-L) 4~(3) (2019)
  3037--3044.
\newblock \href {http://dx.doi.org/10.1109/LRA.2019.2923960}
  {\path{doi:10.1109/LRA.2019.2923960}}.

\bibitem{sun_recurrent-octomap_2018}
L.~Sun, Z.~Yan, A.~Zaganidis, C.~Zhao, T.~Duckett, Recurrent-{OctoMap}:
  Learning state-based map refinement for long-term semantic mapping with
  {3D}-lidar data, IEEE Robotics and Automation Letters (RA-L) 3~(4) (2018)
  3749--3756.
\newblock \href {http://dx.doi.org/10.1109/LRA.2018.2856268}
  {\path{doi:10.1109/LRA.2018.2856268}}.

\bibitem{hornung2013octomap}
A.~Hornung, K.~M. Wurm, M.~Bennewitz, C.~Stachniss, W.~Burgard, {OctoMap}: {An}
  efficient probabilistic {3D} mapping framework based on octrees, Autonomous
  Robots 34~(3) (2013) 189--206.
\newblock \href {http://dx.doi.org/10.1007/s10514-012-9321-0}
  {\path{doi:10.1007/s10514-012-9321-0}}.

\bibitem{landgraf2020comparing}
Z.~Landgraf, F.~Falck, M.~Bloesch, S.~Leutenegger, A.~J. Davison, Comparing
  view-based and map-based semantic labelling in real-time {SLAM}, in: IEEE
  Int. Conf. on Robotics and Automation (ICRA), 2020, pp. 6884--6890.
\newblock \href {http://dx.doi.org/10.1109/ICRA40945.2020.9196843}
  {\path{doi:10.1109/ICRA40945.2020.9196843}}.

\bibitem{mascaro2021diffuser}
R.~Mascaro, L.~Teixeira, M.~Chli, Diffuser: Multi-view {2D}-to-{3D} label
  diffusion for semantic scene segmentation, in: IEEE Int. Conf. on Robotics
  and Automation (ICRA), 2021, pp. 13589--13595.
\newblock \href {http://dx.doi.org/10.1109/ICRA48506.2021.9561801}
  {\path{doi:10.1109/ICRA48506.2021.9561801}}.

\bibitem{berrio2020proj}
J.~S. Berrio, M.~Shan, S.~Worrall, J.~Ward, E.~Nebot, Semantic sensor fusion:
  From camera to sparse lidar information, preprint arXiv:2003.01871.

\bibitem{maturana_looking_2017}
D.~Maturana, S.~Arora, S.~Scherer, Looking forward: {A} semantic mapping system
  for scouting with micro-aerial vehicles, in: IEEE/RSJ Int. Conf. on
  Intelligent Robots and Systems (IROS), 2017, pp. 6691--6698.
\newblock \href {http://dx.doi.org/10.1109/IROS.2017.8206585}
  {\path{doi:10.1109/IROS.2017.8206585}}.

\bibitem{dengler2021ecmr}
N.~Dengler, T.~Zaenker, F.~Verdoja, M.~Bennewitz, Online object-oriented
  semantic mapping and map updating, in: Europ. Conf. on Mobile Robots (ECMR),
  2021, pp. 1--7.
\newblock \href {http://dx.doi.org/10.1109/ECMR50962.2021.9568817}
  {\path{doi:10.1109/ECMR50962.2021.9568817}}.

\bibitem{ren2017faster}
S.~Ren, K.~He, R.~Girshick, J.~Sun, Faster {R-CNN}: Towards real-time object
  detection with region proposal networks, IEEE Trans. on Pattern Analysis and
  Machine Intelligence 39~(6) (2017) 1137--1149.
\newblock \href {http://dx.doi.org/10.1109/TPAMI.2016.2577031}
  {\path{doi:10.1109/TPAMI.2016.2577031}}.

\bibitem{rosu2020semi}
R.~A. Rosu, J.~Quenzel, S.~Behnke, Semi-supervised semantic mapping through
  label propagation with semantic texture meshes, Int. Journal of Computer
  Vision (IJCV) 128~(5) (2020) 1220--1238.
\newblock \href {http://dx.doi.org/10.1007/s11263-019-01187-z}
  {\path{doi:10.1007/s11263-019-01187-z}}.

\bibitem{langer2020lidar_dom_transfer}
F.~Langer, A.~Milioto, A.~Haag, J.~Behley, C.~Stachniss, Domain transfer for
  semantic segmentation of {LiDAR} data using deep neural networks, in:
  IEEE/RSJ Int. Conf. on Intelligent Robots and Systems (IROS), 2020, pp.
  8263--8270.
\newblock \href {http://dx.doi.org/10.1109/IROS45743.2020.9341508}
  {\path{doi:10.1109/IROS45743.2020.9341508}}.

\bibitem{yi2021scvn}
L.~Yi, B.~Gong, T.~Funkhouser, Complete \& label: A domain adaptation approach
  to semantic segmentation of {LiDAR} point clouds, in: IEEE Conf. on Computer
  Vision and Pattern Recognition (CVPR), 2021, pp. 15358--15368.
\newblock \href {http://dx.doi.org/10.1109/CVPR46437.2021.01511}
  {\path{doi:10.1109/CVPR46437.2021.01511}}.

\bibitem{alonso2021lidaradapt}
I.~Alonso, L.~Riazuelo, L.~Montesano, A.~Murillo, Domain adaptation in {LiDAR}
  semantic segmentation by aligning class distributions, in: Int. Conf. on
  Informatics in Control, Automation and Robotics - ICINCO,, 2021, pp.
  330--337.
\newblock \href {http://dx.doi.org/10.5220/0010610703300337}
  {\path{doi:10.5220/0010610703300337}}.

\bibitem{liu2021otoc}
Z.~Liu, X.~Qi, C.-W. Fu, One thing one click: A self-training approach for
  weakly supervised {3D} semantic segmentation, in: IEEE Conf. on Computer
  Vision and Pattern Recognition (CVPR), 2021, pp. 1726--1736.
\newblock \href {http://dx.doi.org/10.1109/CVPR46437.2021.00177}
  {\path{doi:10.1109/CVPR46437.2021.00177}}.

\bibitem{liu2019iccv}
B.~Liu, Z.~Wu, H.~Hu, S.~Lin, Deep metric transfer for label propagation with
  limited annotated data, in: IEEE Int. Conf. on Computer Vision Workshops
  (ICCVW), 2019, pp. 1317--1326.
\newblock \href {http://dx.doi.org/10.1109/ICCVW.2019.00167}
  {\path{doi:10.1109/ICCVW.2019.00167}}.

\bibitem{piewak2018lilanet}
F.~Piewak, P.~Pinggera, M.~Sch{\"a}fer, D.~Peter, B.~Schwarz, N.~Schneider,
  M.~Enzweiler, D.~Pfeiffer, M.~Z{\"o}llner, Boosting {LiDAR}-based semantic
  labeling by cross-modal training data generation, in: Europ. Conf. on
  Computer Vision (ECCV) Workshops, 2018, pp. 497--513.
\newblock \href {http://dx.doi.org/10.1007/978-3-030-11024-6_39}
  {\path{doi:10.1007/978-3-030-11024-6_39}}.

\bibitem{jaritz2019xmuda}
M.~Jaritz, T.-H. Vu, R.~de~Charette, E.~Wirbel, P.~P{\'e}rez, {xMUDA}:
  Cross-modal unsupervised domain adaptation for {3D} semantic segmentation,
  in: IEEE Conf. on Computer Vision and Pattern Recognition (CVPR), 2020, pp.
  12602--12611.
\newblock \href {http://dx.doi.org/10.1109/CVPR42600.2020.01262}
  {\path{doi:10.1109/CVPR42600.2020.01262}}.

\bibitem{wang2021multistage}
Z.~Wang, Z.~Zhao, Z.~Jin, Z.~Che, J.~Tang, C.~Shen, Y.~Peng, Multi-stage fusion
  for multi-class {3D} lidar detection, in: IEEE Int. Conf. on Computer Vision
  Workshops (ICCVW), 2021, pp. 3113--3121.
\newblock \href {http://dx.doi.org/10.1109/ICCVW54120.2021.00347}
  {\path{doi:10.1109/ICCVW54120.2021.00347}}.

\bibitem{MVD2017}
G.~Neuhold, T.~Ollmann, S.~Rota~Bul\`o, P.~Kontschieder, The mapillary vistas
  dataset for semantic understanding of street scenes, in: IEEE Int. Conf. on
  Computer Vision (ICCV), 2017, pp. 5000--5009.
\newblock \href {http://dx.doi.org/10.1109/ICCV.2017.534}
  {\path{doi:10.1109/ICCV.2017.534}}.

\bibitem{lin_coco_2014}
T.-Y. Lin, M.~Maire, S.~Belongie, J.~Hays, P.~Perona, D.~Ramanan,
  P.~Doll{\'a}r, C.~L. Zitnick, {Microsoft COCO: Common objects in context},
  in: Europ. Conf. on Computer Vision (ECCV), 2014, pp. 740--755.
\newblock \href {http://dx.doi.org/10.1007/978-3-319-10602-1_48}
  {\path{doi:10.1007/978-3-319-10602-1_48}}.

\bibitem{flir_dataset}
FLIR, {FLIR} thermal dataset for algorithm training,
  \url{https://www.flir.com/oem/adas/adas-dataset-form}, accessed: 2022-01-10
  (2022).

\bibitem{abadi2016tensorflow}
M.~Abadi, P.~Barham, J.~Chen, Z.~Chen, A.~Davis, J.~Dean, M.~Devin,
  S.~Ghemawat, G.~Irving, M.~Isard, et~al., {TensorFlow}: A system for
  large-scale machine learning, in: USENIX symposium on operating systems
  design and implementation (OSDI), 2016, pp. 265--283.

\bibitem{quantization_2018}
B.~Jacob, S.~Kligys, B.~Chen, M.~Zhu, M.~Tang, A.~Howard, H.~Adam,
  D.~Kalenichenko, {Quantization and training of neural networks for efficient
  integer-arithmetic-only inference}, in: IEEE Conf. on Computer Vision and
  Pattern Recognition (CVPR), 2018, pp. 2704--2713.
\newblock \href {http://dx.doi.org/10.1109/CVPR.2018.00286}
  {\path{doi:10.1109/CVPR.2018.00286}}.

\bibitem{quenzel2021mars}
J.~Quenzel, S.~Behnke, Real-time multi-adaptive-resolution-surfel {6D LiDAR}
  odometry using continuous-time trajectory optimization, in: IEEE/RSJ Int.
  Conf. on Intelligent Robots and Systems (IROS), 2021, pp. 5499--5506.
\newblock \href {http://dx.doi.org/10.1109/IROS51168.2021.9636763}
  {\path{doi:10.1109/IROS51168.2021.9636763}}.

\bibitem{deschaud2021paris}
J.-E. Deschaud, D.~Duque, J.~P. Richa, S.~Velasco-Forero, B.~Marcotegui,
  F.~Goulette, {Paris-CARLA-3D}: A real and synthetic outdoor point cloud
  dataset for challenging tasks in {3D} mapping, Remote Sensing 13~(22) (2021)
  4713.
\newblock \href {http://dx.doi.org/10.3390/rs13224713}
  {\path{doi:10.3390/rs13224713}}.

\bibitem{geiger2012cvpr}
A.~Geiger, P.~Lenz, R.~Urtasun, Are we ready for autonomous driving? the
  {KITTI} vision benchmark suite, in: IEEE Conf. on Computer Vision and Pattern
  Recognition (CVPR), 2012, pp. 3354--3361.
\newblock \href {http://dx.doi.org/10.1109/CVPR.2012.6248074}
  {\path{doi:10.1109/CVPR.2012.6248074}}.

\end{thebibliography}

\end{document}